\documentclass[letterpaper]{article} %

\usepackage{aaai2026}  %
\usepackage{times}  %
\usepackage{helvet}  %
\usepackage{courier}  %
\usepackage[hyphens]{url}  %
\usepackage{graphicx} %
\usepackage{natbib}  %
\usepackage{caption} %
\usepackage{algorithm}
\usepackage{algorithmic}

\usepackage{newfloat}
\usepackage{listings}
\DeclareCaptionStyle{ruled}{labelfont=normalfont,labelsep=colon,strut=off} %
\floatstyle{ruled}
\newfloat{listing}{tb}{lst}{}
\floatname{listing}{Listing}
\usepackage{microtype}
\usepackage{graphicx}
\usepackage{subfigure}
\usepackage{booktabs} %

\usepackage{amsmath}
\usepackage{amssymb}
\usepackage{mathtools}
\usepackage{amsthm}

\theoremstyle{plain}
\newtheorem{theorem}{Theorem}[section]
\newtheorem{proposition}[theorem]{Proposition}

\theoremstyle{definition}

\theoremstyle{remark}

\usepackage[textsize=tiny]{todonotes}

\usepackage{xspace}  %
\usepackage{enumitem}  %
\usepackage{amssymb}
\usepackage{amsbsy}
\usepackage{bbding}

\newcommand{\method}{\textsc{PHE}\xspace}

\usepackage{amsmath,amsfonts,bm}

\def\eqref#1{equation~\ref{#1}}

\def\1{\bm{1}}

\def\rvh{{\mathbf{h}}}
\def\rvu{{\mathbf{i}}}

\def\rvm{{\mathbf{m}}}

\def\rvs{{\mathbf{s}}}

\def\rvu{{\mathbf{u}}}

\def\rvx{{\mathbf{x}}}
\def\rvy{{\mathbf{y}}}
\def\rvz{{\mathbf{z}}}

\DeclareMathAlphabet{\mathsfit}{\encodingdefault}{\sfdefault}{m}{sl}
\SetMathAlphabet{\mathsfit}{bold}{\encodingdefault}{\sfdefault}{bx}{n}

\def\gD{{\mathcal{D}}}

\def\gL{{\mathcal{L}}}

\def\gN{{\mathcal{N}}}

\def\gS{{\mathcal{S}}}

\def\sN{{\mathbb{N}}}

\def\sR{{\mathbb{R}}}

\newcommand{\E}{\mathbb{E}}

\newcommand{\softmax}{\mathrm{softmax}}

\newcommand{\KL}{D_{\mathrm{KL}}}

\DeclareMathOperator*{\argmax}{arg\,max}

\newcommand{\x}{\rvx}
\newcommand{\m}{\rvm}

\newcommand{\z}{\rvz}
\renewcommand{\u}{\rvu}

\newcommand{\s}{\rvs}
\newcommand{\y}{\rvy}
\newcommand{\h}{\rvh}

\usepackage[capitalise,noabbrev]{cleveref}
\Crefname{figure}{Fig.}{Figs.}%
\Crefname{table}{Tab.}{Tabs.}%
\Crefname{section}{Sec.}{Secs.}%
\Crefname{equation}{Eq.}{Eqs.}%
\Crefname{appendix}{Supp.}{Supps.}%
\Crefname{proposition}{Prop.}{Props.}%

\title{Probabilistic Hash Embeddings for Online Learning of Categorical Features}
\author{
    Aodong Li,
    Abishek Sankararaman,
    Balakrishnan Murali Narayanaswamy
}
\affiliations{
    Amazon Web Services\\
    aodongli@amazon.com, abisanka@amazon.com, muralibn@amazon.com
}

\begin{document}

\maketitle

\begin{abstract}

We study streaming data with categorical features where the vocabulary of categorical feature values is changing and can even grow unboundedly over time.
Feature hashing is commonly used as a pre-processing step to map these categorical values into a feature space of fixed size before learning their embeddings. While these methods have been developed and evaluated for offline or batch settings, in this paper we consider  online settings. 
We show that deterministic embeddings are sensitive to the arrival order of categories and suffer from forgetting in online learning, leading to performance deterioration. 
To mitigate this issue, we propose a probabilistic hash embedding (PHE) model that treats hash embeddings as stochastic and applies Bayesian online learning to learn incrementally from data. 
Based on the structure of PHE, we derive a scalable inference algorithm to learn model parameters and infer/update the posteriors of hash embeddings and other latent variables. 
Our algorithm (i) can handle an evolving vocabulary of categorical items, (ii) is adaptive to new items without forgetting old items, (iii) is implementable with a bounded set of parameters that does not grow with the number of distinct observed values on the stream, and (iv) is invariant to the item arrival order. 
Experiments in classification, sequence modeling, and recommendation systems in online learning setups demonstrate the superior performance of PHE while maintaining high memory efficiency (consumes as low as 2$\sim$4\% memory of a one-hot embedding table).

\end{abstract}

\begin{links}
    \link{Code}{https://github.com/aodongli/probabilistic-hash-embeddings}
    \link{Extended version}{https://arxiv.org/abs/2511.20893}
\end{links}

\section{Introduction}
\label{sec:intro}

Categorical features occur in many high-value ML applications:
finance~\citep{clements2020sequential}, fraud detection~\citep{al2021financial}, anomaly detection~\citep{han2022adbench}, cybersecurity~\citep{sarker2020cybersecurity}, medical diagnosis~\citep{shehab2022machine}, recommendation systems~\citep{ko2022survey,lai2023adaembed} etc. 
Large vocabularies and embedding tables are strong characteristics of these categorical feature-intensive applications. While assigning each item\footnote{
  We use the term \emph{item} to refer to a categorical value.
}
its own row in a large embedding table typically improves accuracy, this comes at a cost. Larger embedding tables require more resources and memory to deploy and slow down execution.

A common and well-established solution for learning categorical features at scale, without maintaining a large vocabulary and embedding table, is the use of hashing techniques \citep{weinberger2009feature,tito2017hash,shi2020beyond,kang2021learning,lai2023adaembed,coleman2023unified}. 
For any categorical value (typically a string), a hash function maps it to an index in a small, fixed range. 
This index addresses a row in a much smaller embedding table. These rows, referred to as \textit{hash embeddings}, are used in subsequent model training and inference.

Hash collisions--when different items map to the same hashed value and thus share an embedding--can degrade model performance. This issue can be mitigated in two ways: first, by designing sophisticated operations on the hashed values~\citep{weinberger2009feature}; second, by using multiple hash functions~\citep{tito2017hash,coleman2023unified}. Large technology firms like Yahoo and Google have successfully incorporated this approach in their applications~\citep{weinberger2009feature,coleman2023unified}.

A major limitation of prior works is they focused on \emph{offline} settings where data distributions are stationary and the entire test set is available in batch. In other words, the vocabulary of categorical items is fixed with no new or out-of-vocabulary items occurring during testing.

\begin{figure}[t]
\centering
           \includegraphics[width=0.35\textwidth]{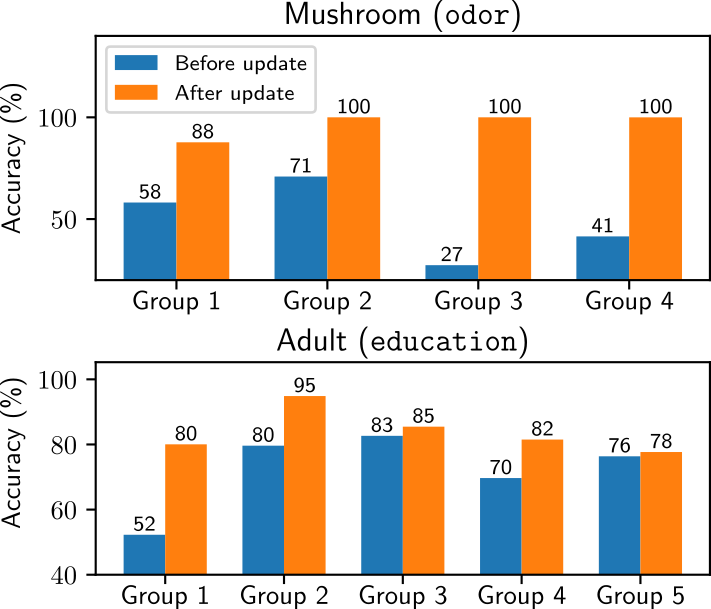}
        \caption{On two tabular datasets, Mushroom and Adult,
        we split the data into groups based on random partitions of categorical column vocabularies, such that each group has disjoint vocabulary sets. We report the results before and after online learning on each  group. The performance gaps motivate the need to learn representations of new items. Bracketed values indicate the splitting columns. 
        Results are averaged over five runs. 
        Partition details are in \Cref{fig:group-info}.
            }
        \label{fig:group-update-perf}
\end{figure}

In many real applications, data arrives in a streaming fashion:
{\em(i)} the vocabulary of categorical items can change, and/or {\em(ii)} the meaning of an item can evolve.
These characteristics present challenges to offline predictive models. 
Failure to adapt to the expanding vocabulary leads to a loss in predictive performance, as shown in Fig. \ref{fig:group-update-perf}--in streaming binary classifications, new groups or sets of categorical items occur sequentially, and modeling these new items incrementally leads to significant accuracy improvement.
This problem of vocabulary expansion is fairly common in practice: new products are added to grocery stores \citep{cheng2023efficient}, new usernames and application names appear in intrusion detection systems \citep{siadati2017detecting,le2022rethinking}, new patients arrive at hospitals, and so on. 

More critically, as our analysis and experiments show (\Cref{sec:minor_theory,sec:exp}), naively adapting to new items while using hashing techniques suffers from \textit{catastrophic forgetting} \citep{kirkpatrick2017overcoming}. In hash embeddings, different items may share embeddings, updating one can adversely affect another, causing the model to effectively ``forget.'' Furthermore, this forgetting can be exacerbated depending on the arrival order.
Consequently, hash embeddings are not yet fit for online learning in their vanilla form.

In this paper, we employ Bayesian online learning to mitigate the forgetting issue. This approach is theoretically shown to be as effective as offline batch learning, regardless of the order in which items arrive~\citep{opper1999bayesian}. Specifically, we model hash embeddings as random variables and perform posterior inference upon observing new data. This probabilistic treatment enables continual adaptation to new items while preserving knowledge of previously seen ones.

\textbf{Main Contributions:}
Our work proposes \textit{probabilistic hash embeddings} (PHE) with Bayesian updates to handle dynamic vocabularies effectively and efficiently. 
The intuition behind PHE stems from its benefits in {\em(i)} efficiency, as memory and the number of model parameters is bounded, with only a small number of parameters requiring online updates (ie., less forgetting), and {\em(ii)} accuracy, as the Bayesian treatment provides implicit regularization that balances forgetting and adaptation while maintaining invariance to item arrival order, without  dataset-specific regularization design.

We highlight PHE as a plug-in module, which can be applied to other probabilistic models such as Deep Kalman Filters \citep{krishnan2015deep}
and Neural Collaborative Filtering \citep{he2017neural}.
The usage of PHE allows these models to handle unbounded items in their domains in a principled way.
We derive scalable variational inference algorithms for learning PHE and provide theoretical analysis demonstrating PHE's superiority for online learning.
Empirically, our method outperforms baselines in three setups: online supervised learning with new items, conditional sequence modeling with growing vocabularies, and a recommendation system with evolving user-item interactions.

\textbf{Organization:} We survey  related work  in \Cref{sec:background}, present \method, derive its inference algorithms in \Cref{sec:meth},  demonstrate \method's efficacy in \Cref{sec:exp}, and conclude in \Cref{sec:concl}. More details and limitation discussions are in supplementary materials.

\section{Related Work}
\label{sec:background}

\paragraph{Hashing trick.} \citet{weinberger2009feature} first proposed using hashing to handle an unbounded number of categorical items. To mitigate degradation due to hash collisions, \citet{serra2017getting} used bloom filters. More recently, \citet{tito2017hash,cheng2023efficient, coleman2023unified} proposes shared embeddings across all categorical features for efficiency and multiple hashing functions to reduce collisions. However, unlike our method, these approaches are deterministic and are developed in offline settings. As shown in our experiments, deterministic hashing embeddings are vulnerable to evolving vocabularies.

\paragraph{Continual learning.} 
Previous continual learning methods \citep{kirkpatrick2017overcoming,nguyen2018variational,lopez2017gradient} were designed for continuous-valued features and do not address how to learn evolving categorical features.
This only existing approach, architecture-expanding methods that dynamically expand the embedding tables (see EE in experiments), leads to unbounded memory usage~\citep{rusu2016progressive, yoon2017lifelong,jerfel2019reconciling}. In contrast, \method is the first to maintain constant memory while preserving accuracy and invariance to category arrival order.

\paragraph{Temporal and recommendation models.} Temporal and recommendation models are important applications for categorical feature embeddings. We extend Deep Kalman Filters (DKF)~\citep{krishnan2015deep} to handle evolving vocabularies, whereas the original DKF assumes a fixed vocabulary.  
Similarly, previous recommendation methods \citep{ko2022survey,shi2020beyond,kang2021learning,coleman2023unified} assume training data is given in batch and the item vocabulary remains fixed. 

\paragraph{Tabular data models.} 
Categorical features are ubiquitous in tabular datasets. 
In online and continual learning settings, deep learning-based methods have been studied in recent years \citep{huang2020tabtransformer,du2021tabularnet,liu2023incremental}. However, all of these works assume the vocabulary of categorical items is known and fixed in advance. {\em Ours is the first online learning method for tabular data that can handle increasing and unbounded items.} 
While \citet{kimcarte} use string embeddings from language models for open-vocabulary categorical features in an offline setting, we focus on online settings. Moreover, language model representations cannot be exploited when strings lack semantic meaning (such as anonymized ZIP codes, IP addresses, etc.). We survey additional related work in \Cref{app:related}.

\section{Methodology}
\label{sec:meth}
\begin{figure}
    \centering
           \includegraphics[width=0.4\textwidth]{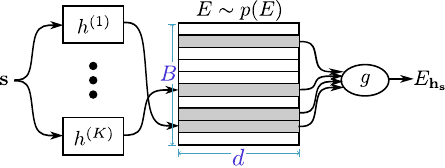}
        \caption{PHE architecture.
        The input $\rvs$ can be usernames or anonymized strings. 
        The entire module represents $p(E_{\h_\s})$.}
         \label{fig:hash-emb}
\end{figure}

We first establish the necessary notations and introduce our proposed probabilistic hash embedding (PHE) module. Next, we derive an online inference algorithm for PHE. Finally, we analyze and explain why PHE is superior in online learning.

\subsection{Notations and Problem Setup}
We consider the problem of learning predictive ML models with categorical features in an online learning environment. We assume the model input and output dimensions are fixed but new categorical features may occur over time.\footnote{
  We also assume categorical features are single-valued, that is, each feature is assigned exactly one category. However, our work is compatible with multi-valued features.
}

We denote categorical items by $s\in \mathcal{S}$. Typically, practitioners aim to predict certain target features using ML models based on other input features. We denote these target features by $y$. Depending on the task, $y$ can be either numerical or categorical. We use subscript $i$ to denote the $i$-th data point.

Hashing techniques use hash functions (e.g., MD5) to map categorical items to hashed values. These hashed values are then used to index rows of an embedding table. The range of hash values is usually much smaller than the vocabulary size of the categorical items, thus achieving memory and computational efficiency. We define a hash function $h:\mathcal{S}\rightarrow \mathbb{N}_{<B}$  that maps an item to a hash value, where $B$ is a pre-defined upper bound known as the ``bucket size.''
For simplicity, we use $h_s$ to denote the hash value $h(s)$ of an item $s$. 
The hash value $h_s$ indexes a row in the hash embedding table $E\in\mathbb{R}^{B\times d}$, yielding the embedding representation $E_{h_s}$ of $s$.
We use the same notation for both random variables and their realizations when the meaning is clear from context.

\subsection{Probabilistic Hash Embeddings (PHE)}
\label{sec:met-hash-emb}
We first introduce our proposed encoding module for categorical items--probabilistic hash embeddings (PHE). A basic PHE has two components: a fixed hash function $h$ and a hash embedding table $E$ with a prior distribution $p(E)$. In this work, we assume each entry of $E$ is independently Gaussian-distributed. Given an item $s$, PHE  looks up the $h_s$-th row of $E$ as its embedding $E_{h_s}$, whose distribution is $p(E_{h_s})$.

A single hash function may result in distinct inputs having the same hashing value, known as a hash collision, resulting in indistinguishable hash embeddings. 
With bucket size $B$, the collision probability is $O(1/B)$. To further reduce the collision rate, we employ universal hashing~\citep{carter1977universal}. 
Namely, instead of using one hash function, we use $K$ hash functions with different random seeds but the same range. 
The collision probability then reduces to $O(1/B^K)$. Moreover, we share the same hash embedding table $E$ across  all $K$ hash functions, which keeps the model size bounded. 

We now describe how PHE encodes an item $s$. This procedure is illustrated in \Cref{fig:hash-emb}. %
With $K$ hash functions, a categorical feature $s$ results in $K$ hash values $\h_s:=\{h^{(1)}_s,\dots,h^{(K)}_s\}$, which index $K$ embeddings 
$\{E_{h^{(1)}_s},\dots,E_{h^{(K)}_s}\}$
where $E_{h^{(k)}_s}$ is the embedding retrieved using the $k$-th hash value $h^{(k)}_s$. The final representation $E_{\h_s}:=g(E_{h^{(1)}_s},\dots,E_{h^{(K)}_s})$ of $s$ is obtained using an assembly function $g:\sR^{K\times d}\rightarrow\sR^d$ to combine the $K$ embeddings. 
Typical choices of $g$ involve coordinate-wise summation, average, maximum, or minimum; other parametric choices of $g$ include weighted sums where the weights come from another parametric model. The output of PHE is a probabilistic embedding $E_{\h_s}$ with distribution $p(E_{\h_s})$.
Due to the shared embedding table, the memory cost of PHE is $O(Bd)$, independent of the number of hash functions $K$.

One core ML task is to model correlations between two variables.
A common query is to compute the probability of observing a value of $y$ given a categorical item $s$, i.e., $p(y|s)$. With PHE, we can compute it as
\begin{align}
\label{eq:cate-cond}
    p(y|s) = \E_{p(E)}[p(y|E, \h_s)]  = \E_{p(E)}[p(y|E_{\h_s})]
\end{align}
where we follow the data generating process and treat $\h_s$ to be the same as $s$ (which holds in the absence of hash collisions). To compute \Cref{eq:cate-cond}, we use Monte Carlo sampling to estimate the expectation. (See \Cref{app:trade-off} for practical implementation details.)
This approach enables probabilistic inference conditioned on categorical features.

\subsection{Online Learning of PHE}
\label{sec:meth-data-gen}

In this subsection, we derive scalable inference algorithms for PHE in a simple yet general model. These algorithms can be adapted to other model variants. We first focus on the static setting before expanding to the online setting.

Given observations $\mathcal{D}:=\{(y_i, \s_{i})\}_{i=1}^N$ where $\s_i$ represents a set of categorical features, we want to learn correlations between $y$ and $\s$ using PHE. We assume observations are conditionally independently and identically distributed (i.i.d.) given $E$, with likelihood $p(y|E_{\h_\s})$.\footnote{
  We use $E_{\h_\s}$ to denote the concatenation of hash embeddings for each item if $\s$ contains more than one categorical items.
}
PHE places a prior $p(E)$ over the hash embedding table $E$ and infers the posterior given the observations $\mathcal{D}$. By Bayes rule, the posterior is $p(E|\mathcal{D})\propto p(E)p(\gD|E)=p(E)\prod_{i=1}^N p(y_i|E_{\h_{\s_i}})$, which is typically intractable with complex likelihoods. %

Therefore, we turn to variational inference \citep{blei2017variational,zhang2018advances} and learn an approximate posterior by minimizing the Kullback-Leibler (KL) divergence $\KL(q_\lambda(E)|p(E|\mathcal{D}))$ between a variational distribution and the true posterior. We assume the variational posterior factorizes as $q_{\lambda}(E) = \prod_{b=1}^B\prod_{j=1}^d q_{\lambda_{bj}}(E_{bj})$ where each factor $q_{\lambda_{bj}}(E_{bj})$ is Gaussian with parameters $\lambda_{bj}:=\{\mu_{bj}\in\mathbb{R}, \sigma_{bj}\in\mathbb{R}\}$. 
Finding the optimal variational distribution that minimizes the KL divergence is equivalent to finding the optimal parameters $\lambda^*:=\{\lambda_{bj}^*\}$.

We assume the prior has the same factorization, with each entry independently following a standard Gaussian, i.e., $p(E_{bj})=\mathcal{N}(0,1)$. Substituting $p(E)$ and $q_\lambda(E)$ into the KL divergence yields an objective function $\mathcal{L}(\lambda)$ (also known as the evidence lower bound, ELBO) to be maximized (see derivations in \Cref{app:elbo}):
\begin{multline}
    \textstyle
    \mathcal{L}(\lambda):= \E_{q_\lambda(E)}\left[\sum_{i=1}^N\log p(y_i|E_{\h_{s_i}})\right] \\
    \textstyle
     -\sum_{b=1}^B\sum_{j=1}^d\KL(q_{\lambda_{bj}}(E_{bj})|p(E_{bj})),\label{eq:elbo-sec3.3}
\end{multline}
where the KL divergence between two Gaussians have a closed-form solution. The KL term also serves as regularization, preventing the variational posterior from deviating too far from the prior. We optimize \Cref{eq:elbo-sec3.3} using the reparametrization trick~\citep{rezende2014stochastic,kingma2013auto} and gradient-based methods. Let $\lambda_0^*:=\argmax \gL(\lambda)$ denote the optimal parameters, which define the approximate posterior $q_{\lambda_0^*}(E)$ that minimizes $\KL(q_{\lambda_0^*}(E)|P(E|\gD))$. For prediction on an unseen data point $(\hat y, \hat s)$, the predictive distribution is $p(\hat y|\hat s,\gD)=\E_{p(E|\gD)}[p(\hat y|E_{\h_{\hat s}})]\approx \E_{q_{\lambda_0^*}(E)}[p(\hat y|E_{\h_{\hat s}})]$.

Suppose we observe a second dataset $\gD_1:=\{(y_i, \s_i)\}_{i=1}^{N_1}$ to which we would like our model to adapt while maintaining effectiveness on $\gD$.
Bayes rule provides a principled online learning approach: $p(E|\gD_1, \gD)\propto p(E|\gD)p(\gD_1|E)$. This allows us to accommodate both datasets without storing them simultaneous. We iteratively replace the prior with the previous posterior and repeat the inference procedure. This iteration assumes $\mathcal{D}$ and $\mathcal{D}_1$ are conditionally i.i.d. given $E$.

Variational inference produces an approximation of the true posterior. We thus use $q_{\lambda_0^*}(E)$ in place of $p(E|\gD)$ and infer $\tilde p(E|\gD_1, \gD)\propto q_{\lambda_0^*}(E)p(\gD_1|E)$. This again faces intractability, so we apply variational inference with a new variational distribution $q_\lambda(E)$ to minimize $\KL(q_\lambda(E)|\tilde p(E|\gD_1, \gD))$.
Rewrite the KL divergence gives us
a new ELBO (the objective function)
\begin{multline}
    \textstyle
    \gL^{(1)}(\lambda;\lambda_0^*) 
    := \E_{q_\lambda(E)}\left[\sum_{i=1}^{N_1}\log p(y_i|E_{\h_{s_i}})\right] \\
    \textstyle
    -\sum_{b=1}^B\sum_{j=1}^d\KL(q_{\lambda_{bj}}(E_{bj})|q_{\lambda_{0,bj}^*}(E_{bj}))\label{eq:elbo-online-sec3.3}.
\end{multline}
Compared to \Cref{eq:elbo-sec3.3}, the original prior $p(E)$ of $E$ is replaced with the previous posterior approximation $q_{\lambda^*_0}(E)$. 
Upon optimization convergence, the new optimal variational distribution approximates the posterior given both datasets ($\gD$ and $\gD_1$).
When new datasets arrive in the future, we repeat the above online learning iteration to accommodate new datasets.
Although this procedure resembles traditional continual learning~\citep{wang2023comprehensive,nguyen2018variational,li2021detecting}, it differs in focusing on evolving discrete vocabularies rather than continuous feature distributions. For any specified ML task, we can plug the corresponding likelihood model into \Cref{eq:elbo-sec3.3,eq:elbo-online-sec3.3} and optimize.

\paragraph{Variational EM.} Most expressive models contain learnable parameters $\theta$ in their likelihood $p_\theta(y|E_{\h_s})$. One needs to learn $\theta$ in addition to inferring the posterior of $E$. Fortunately, a minor modification to our previously derived objective functions (\Cref{eq:elbo-sec3.3,eq:elbo-online-sec3.3}) can achieve this. We replace $p(y|E_{\h_{s}})$ with the parametric one $p_\theta(y|E_{\h_s})$ and maximize $\gL(\lambda,\theta)$ with respect to $\{\lambda,\theta\}$ jointly. These new objective functions are viable and correspond to variational EM algorithms \citep{rezende2014stochastic,kingma2013auto}, for which we provide proofs in \Cref{app:var-em-proof}. Let $\{\lambda^*_0,\theta^*\}$ denote the optimizers of the modified \Cref{eq:elbo-sec3.3}, $\gL(\lambda,\theta)$. During online learning, we fix both $\theta^*$ and $\lambda_0^*$ in the modified objective function \Cref{eq:elbo-online-sec3.3}, $\gL^{(1)}(\lambda;\lambda_0^*,\theta^*)$, to efficiently update the posterior over hash embeddings as new categorical items arrive.

\paragraph{Benefits of PHE in online learning.} In data streaming or continual learning setup, \method has natural benefits in reducing catastrophic forgetting: 1) only a few embeddings need to be updated online. This sparse updating scheme seldom affects other item representations, thus having less forgetting and more computing efficiency. 2) The online updates apply Bayesian online learning, in which the prior distribution serves as a regularization of previous knowledge that also reduces forgetting. Although continually expanding the embedding table with rows for new items typically improves accuracy, it comes at a cost. A larger embedding table requires more resources to deploy, reduces memory efficiency, and slows down execution. 
In contrast, \method's memory/storage cost is bounded and does not increase with the number of distinct categorical values.

\subsection{Why Is PHE Superior for Online Learning?}
\label{sec:minor_theory}

We showcase the benefits of PHE with Bayesian online learning: 1) It is equivalent to batch learning where all data are available at once. 2) This equivalence is independent of the data arrival order.

Given a dataset $\gD=\{(y_i,s_i)\}_{i=1}^N$ and a prior over $E$. Assume data points are conditionally i.i.d. and have likelihood $\prod_{i=1}^N p(y_i|E,s_i)$. The posterior obtained from Bayesian \textit{batch} learning (where we assume the whole dataset is available at once) is $p_\text{batch}(E|\gD)\propto p(E)\prod_{i=1}^N p(y_i|E,s_i)$.

Now, suppose $\boldsymbol{\pi}\!=\!(\pi_1,\dots,\pi_N)$ is any permutation of the sequence $(1,\dots,N)$, and let data in $\gD$ arrive sequentially per the order $\boldsymbol{\pi}$ in an online learning setup. Let $\gD_{\boldsymbol{\pi}}$ denote the dataset that has the arrival order of $\boldsymbol{\pi}$. Suppose the posterior obtained by Bayesian \textit{online} learning is $p(E|\gD_{\boldsymbol{\pi}})$.

\begin{proposition}
\label{prop:bayes-online-learning}
    For every permutation $\boldsymbol{\pi}$, the posterior $p_\text{batch}(E|\gD) = p(E|\gD_{\boldsymbol{\pi}})$ almost-everywhere.
\end{proposition}
We provide the proof in \Cref{app:proof-prop3.1}. Prop.~\ref{prop:bayes-online-learning} shows Bayesian online learning has the same power as Bayesian batch learning irrespective of the data arrival order.

However, for point estimation, such as maximum likelihood estimation, in order to achieve the same optimality as offline batch learning, an online learning algorithm has to be carefully designed and has to have a sophisticated learning rate schedule \citep{opper1999bayesian,orabona2019modern}. 

\paragraph{Demo.} To demonstrate Prop.~\ref{prop:bayes-online-learning}, we design a simple online learning experiment involving two categorical items that arrive alternately. Each item is repeated ten times before the other one arrives. Each item is associated with a target value, and we use both traditional deterministic hash embeddings and PHE to learn these two values in an online fashion. We plot the results in \Cref{fig:synthetic-data-forget} in \Cref{sec:theory}. The results show that while deterministic hash embeddings with a popular online learning algorithm incur large prediction errors on recurrent items, PHE makes much more accurate predictions, corroborating Prop.~\ref{prop:bayes-online-learning}. Detailed settings and error analysis in \Cref{sec:theory} show that the \textit{forgetting} phenomenon seen in traditional hash embeddings is caused by \textit{parameter interference}, and that PHE alleviates the forgetting issue through adaptive regularization of the updated beliefs of hash embeddings.

\paragraph{Practical implementation} Prop.~\ref{prop:bayes-online-learning} holds true for exact Bayesian inference, while our real-world data experiments utilize variational inference (VI) for approximate inference. There are approximation gaps (e.g., mean-field approximation), optimization gaps, and amortization gaps between VI and exact inference. In fact, we use Bayesian inference as a guide, and our experiments conform to the expectations. Secondly, unlike Bayesian neural networks that take long time to converge, PHE has sparse gradient updates and converges much faster. This is because PHE can be set to update only the embedding module while freezing other network parameters. Since each item activates at most $K$ embeddings, by assuming independent embedding slots, the gradient updates on the embedding table are essentially sparse and data efficient. The elapsed time can be found in Sec. F.2.

\section{Experiments}
\label{sec:exp}

In this section, we conduct experiments to demonstrate the efficacy and memory efficiency of \method in online learning settings where categorical features change. 
As follows, we introduce experimental protocols in \Cref{sec:exp-protocol}.
We then showcase the broad applicability of PHE as a plug-in module for a spectrum of models - both deterministic and probabilistic - and benchmark it against baselines in classification, multi-task sequence modeling, and recommendation systems in \Cref{sec:exp-dy-tab,sec:exp-dy-temp-tab,sec:exp-recommend}.
Additional experiments and experimental settings are put in \Cref{sec:exp-add,app:exp}.
All results show that PHE outperforms its deterministic counterpart and performs similarly to the upper-bound collision-free embeddings in various domains and applications.

\subsection{Experimental Protocols and Baselines}
\label{sec:exp-protocol}
\paragraph{Experimental settings.} 
We conducted our experiments using public datasets that contain categorical features and simulated them in data streaming environments where data points arrive sequentially. Our goal is to mimic practical settings where 1) new feature items can emerge and need to be learned, and 2) seen items can recur and need to be remembered.
 
Upon each data arrival, we conducted three operations in order: predict, evaluate, and update (embeddings). We report sequential results in plots and overall averaged results in tables. 
We repeated all experiments five times with different parameter initialization while keeping other settings fixed. 

\paragraph{Baselines.} 
Our work is the first work to handle open vocabulary in online dynamic environments. Appropriate baselines require handling both unbounded feature items and online updates simultaneously. To construct baselines for our setup, we need to combine existing embedding methods designed for offline inference with online updating techniques.

For embedding baselines, we use both hash embeddings (Ada) and collision-free expandable embeddings (EE). While EE is not desirable in practice due to its unbounded memory requirements and slow-down execution, we include it as a baseline to understand the performance gap (if any) resulting from our method's memory efficiency. We also implement a probabilistic version of EE, which we refer to as P-EE, to mitigate potential overfitting through Bayesian treatment.

For Ada, we choose fine-tuning as the online learning strategy, but vary the hyperparameter-training epoch-to give many candidates.
We select different training epochs to account for various distribution shifts: FastAda assumes shifts are rapid while SlowAda assumes shifts happen slowly, and MedAda is in between. At each time step, FastAda trains 15 epochs as EE, while Med/SlowAda reduce training epochs to five and one, respectively. For all Ada baselines, we use the same learning rate as EE. We maintain this protocol across all applications to highlight the robustness of our method. 

\paragraph{Training protocols.} For all methods, we search hyperparameters on a validation set for each application. We test three training epochs (Fast/Medium/Slow) for Ada baselines and report the best results, giving them a competitive advantage. In all experiments, we employ a single shared hash embedding table \citep{coleman2023unified} for all categorical features. During online learning, we update only the hash embedding table while freezing all other model parameters, which were learned on an initial dataset.
To determine the hash embedding table size, We selected $K=3$ and a prime number $B$ to support 10 times the expected size of the total vocabulary (where all tabular datasets use $B=7$, the Retail dataset uses $B=109$, and the MovieLens dataset uses $B=10009$).

\subsection{Application 1: Classification}
\label{sec:exp-dy-tab}

\begin{table*}[t]
\caption{Online learning results on all datasets. Adult, Bank, Mushroom, and Covertype are classification tasks evaluated by average accuracy, the larger the better. Retail and MovieLens-32M use mean absolute error, lower the better. All results are multiplied by 100 except Retail for visual clarity.
PHE achieves the best performance among all hash embedding-based methods.
We also include the compression ratios of PHE with respect to P-EE, which are computed by dividing the number of embedding parameters of PHE by the one of P-EE. (See details in \Cref{tab:memory-cost} and \Cref{sec:app-memory}.) Notably, PHE takes as low as 2\% memory of P-EE.
} \label{tab:cum-res-dy-tab}
\centering
\small
\resizebox{0.91\textwidth}{!}{
\begin{tabular}{l|ccccc|cc}
\toprule
  & \multicolumn{5}{c|}{Hash Embedding} & \multicolumn{2}{|c}{Collision-Free Embedding} \\
\midrule
  &SlowAda &MediumAda &FastAda &PHE (ours) &Compression Ratio of PHE &EE  &P-EE \\
\midrule
Adult ($\uparrow$) 
    & 82.2 $\pm$ 0.7
    & 74.8 $\pm$ 4.5
    & 71.1 $\pm$ 4.0
    & \textbf{84.1 $\pm$ 0.2} &0.09
    & 84.2 $\pm$ 0.0
    & 84.8 $\pm$ 0.0\\
Bank ($\uparrow$) 
    & \textbf{89.7 $\pm$ 0.1}
    & 89.0 $\pm$ 0.9
    & 86.9 $\pm$ 1.6
    & \textbf{89.6 $\pm$ 0.0} &0.2
    & 90.0 $\pm$ 0.0
    & 90.1 $\pm$ 0.0\\
Mushroom ($\uparrow$) 
    & 97.7 $\pm$ 0.7
    & 97.9 $\pm$ 0.5
    & 98.3 $\pm$ 0.3
    & \textbf{98.8 $\pm$ 0.0} &0.62
    & 98.8 $\pm$ 0.0
    & 98.8 $\pm$ 0.0\\
CoverType ($\uparrow$) 
    & 63.5 $\pm$ 0.5
    & 59.1 $\pm$ 1.2
    & 55.3 $\pm$ 1.2
    & \textbf{64.3 $\pm$ 0.2} &0.2
    & 64.3 $\pm$ 0.1
    & 64.0 $\pm$ 0.4\\
Retail ($\downarrow$) 
    &49.1$\pm$82.9
    &22.7$\pm$20.3
    &-  
    &\textbf{3.0$\pm$0.2} &0.02
    &3.7$\pm$0.1 
    &3.2$\pm$0.4\\
MovieLens ($\downarrow$) %
    &15.3$\pm$0.1
    &15.1$\pm$0.1
    &15.1$\pm$0.1
    &\textbf{14.7$\pm$0.0} &0.04
    &15.1$\pm$0.0
    &14.7$\pm$0.0\\
\bottomrule
\end{tabular}
}
\end{table*}

\begin{figure*}
    \centering

    \includegraphics[width=0.95\linewidth]{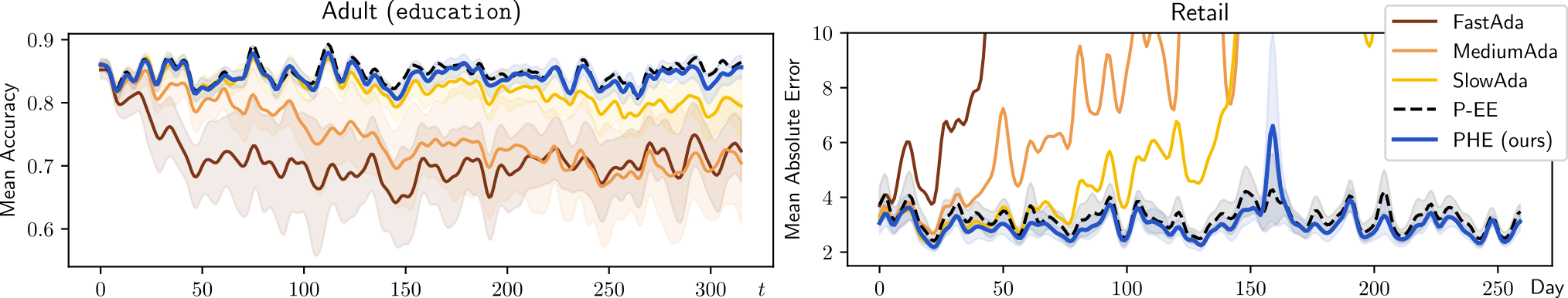}
    \caption{(\textbf{left}) Online classification results on \texttt{Adult} tabular data streams.
    In the parentheses is the column whose items embeddings get updated.
    The Ada results show a downward trend although there are no new items to learn, suggesting the deterministic hash embeddings suffer from forgetting during the learning.  
    In contrast, the proposed PHE mitigates the forgetting issue and keeps performing as good as the upper-bound method P-EE.
    Other datasets in \Cref{fig:all-tab-online} in \Cref{sec:app-classification} show similar conclusions.
    (\textbf{right}) Results of sequence modeling on \texttt{Retail} data-streams. It shows that PHE outperforms all Ada baselines  that are sensitive to their optimization hyperparameters.
    Moreover, it is remarkable to note that PHE performs slightly better than the collision-free P-EE baseline, especially considering PHE consumes only $2\%$ of the memory of P-EE.
    }
    \label{fig:tab-online}
\end{figure*}

We apply PHE to online learning classification tasks on four public datasets from scientific domains. %

\paragraph{Methods.} We now specify the likelihood model for classification tasks. Throughout the four datasets in use, we assume classification target variables follow categorical distributions and have a likelihood function $p_\theta(y|\s,\x)=\text{Cat}(K, \softmax(\theta^\top [\x, E_{\h_\s}]))$ where $\x$ are numeric features.
This likelihood is then plugged into \Cref{eq:elbo-sec3.3,eq:elbo-online-sec3.3}.

\paragraph{Datasets.} 
We apply four public static tabular datasets that are available in UCI Machine Learning Repository: Adult, Bank, Mushroom, and Covertype.
These datasets contain a mixture of discrete and continuous columns and are collected for classification problems in various domains. 
For training stability, we normalized all continuous columns. %

\paragraph{Experimental setups.} 
To simulate the data-streaming setup, at each step we present a randomly sampled data mini-batch to the model and evaluate the online learning performance. 
We require only one column's item embeddings be updated, mimicking that column has a changing vocabulary.
Besides, we initialize the model (both embeddings and neural network weights) with a separate random portion of the data.

\paragraph{Results.}
We reported the data-streaming online classification accuracy in \Cref{fig:tab-online} (left) and \Cref{fig:all-tab-online} in supplement.
The facts that 1) any items seen during online learning have been learned at the initialization and that 2) the accuracy curves of Ada methods have a downward trend suggest hash embeddings suffers from forgetting.
In fact, the forgetting is caused by \textit{parameter interference} in shared hash embeddings: suppose items A and B share parameters in the hash embedding table, then updating A's embedding affect B's embedding. 

We further reported an overall averaged accuracy in \Cref{tab:cum-res-dy-tab}.
The results show that our proposed PHE performs similarly with the upper-bound collision-free embeddings (EE), and the gap between PHE and all other deterministic counterparts proves the effectiveness of PHE in online learning.
Besides, PHE is more stable and has a smaller variance.
Notably, PHE applies the same set of hyperparameters and outperforms all Ada baselines across all datasets. 
The varying performances of the Ada baselines highlight the importance and sensitivity of hyperparameter tuning for deterministic hash embeddings. In contrast, the only demand of our method is to train the model until convergence--a simpler optimization criterion. Lastly, as summarized in \Cref{tab:cum-res-dy-tab,tab:memory-cost}, PHE consumes noticeably lower memory than P-EE.

\subsection{Application 2: Multi-Task Sequence Modeling}
\label{sec:exp-dy-temp-tab}

Sequence models can exploit temporal correlations among observations to make predictions based on histories. 
We now switch to a more sophisticated multi-task sequence modeling problem where each task has its own sequential characteristics, and we aim to personalize the sequence model for each task. Sequential Tasks are identified by categorical features. 

\paragraph{Methods.} We model sequences using deep Kalman filters (DKF) \citep{krishnan2015deep}, which are latent variable models and can handle uncertainties and non-stationary processes. In sequence data, we have an additional timestamp feature $t$. We model the dependency between neighboring data points by a latent time variable $\z$. Specifically, we assume $\z$ is Gaussian distributed and follows the distribution $p(\z_i|\z_{i-1},\Delta_i;\theta_z)=\gN(\z_i|f_{\theta_z}(\z_{i-1}, \Delta_i))$ where $f_{\theta_z}:=\{\mu_{\theta_z}, \Sigma_{\theta_z}\}$ is a multi-layer perceptron that outputs mean and covariance of $\z_i$. $\Delta_i$ is the difference in timestamp between the $i$-th and $i-1$-th observations. 
We assume the conditional likelihood model is $p(y_i|\x_i,E_{\h_{\s_i}},\z_i)$, which is a parametric Poisson distribution.

In the above model, we need to infer the posteriors of $E$ and $\z$. We assume the variational posterior distribution factorizes as
    $q_{\lambda,\phi}(E,\z_{\leq N}|\x_{\leq N}, \y_{\leq N}, \h_{\s_{\leq N}})
    = q_\lambda(E)\prod_{i=1}^N q_\phi(\z_i|\x_{\leq i}, \y_{\leq i}, E_{\h_{\s_{\leq i}}})$
where $q_\lambda(E)$ is the same as before and $q_\phi(\z_i|\x_{\leq i}, \y_{\leq i}, E_{\h_{\s_{\leq i}}})$ is a Gaussian with diagonal covariance implemented as a recurrent neural network with parameters $\phi$. Concretely, we use gated recurrent unit (GRU) \citep{chung2014empirical}. We repeat the KL divergence minimization derivation procedure to obtain the following ELBO objective function
$\gL(\theta, \lambda, \phi)
:=\E_{q_\lambda(E)}\left[\sum_{i=1}^N\gL_i(\theta, \phi|E)\right] - \KL(q_\lambda(E)|p(E))$
where $\gL_i(\theta,\phi|E)$ is the conditional ELBO of the $i$-th data's log-likelihood 
$\gL_i(\theta,\phi|E)
    :=\E_{q_\phi(\z_i)}[\log p(\y_i|\z_i,\x_i,E_{\h_{\s_i}};\theta_y)] 
     - \E_{q_\phi(\z_{i-1})}[\KL(q_\phi(\z_i)|p(\z_i|\z_{i-1};\theta_z))]$.
We provide the full derivation of these ELBOs in \Cref{app:elbo}. After learning the initial dataset, we will fix all model parameters except the hash embedding table $E$ in learning future datasets.

\paragraph{Datasets.}
We apply a public large-scale time-stamped tabular dataset, Retail. A snippet of this dataset can be found in \Cref{tab:retail-snippet}.
This dataset records all online transactions between 01/12/2010 and 09/12/2011 in a retail store. 
There are over 4,000 products and over 540K time-stamped invoice records in total.
The task is to predict the sales for each product shown in each invoice given the product's historical sales. 

\paragraph{Experimental setups.}
We use the first three month data to initialize the model. Then we make predictions on a daily basis following the invoice timestamp. 
And at each step, we predict the sales quantity for each product on invoices based on their sale history.
After that, we will receive the prediction error and use it to update the product embeddings.
We use mean absolute errors for evaluation (see \Cref{sec:app-sequence-mdoel}).

\paragraph{Results.}
\Cref{fig:tab-online} (right) shows the running performance (smoothed by a 1-D Gaussian filter): the Ada-family baselines favor shorter optimization time for Retail, as long optimization time like FastAda explodes after 50 days.
(The error bar is omitted as it is too large to be meaningful.)
On the other hand, \method has lower error and is stable across all learning steps. 
Remarkably, on the average performance in \Cref{tab:cum-res-dy-tab}, PHE significantly outperforms all baselines, including collision-free P-EE with only 2\% memory usage. One possible reason is that P-EE initializes new embeddings from scratch and thus gets slow in warm-up, while PHE uses shared parameters from initial training.
Similar observations also occur in the continual learning setup (see \Cref{fig:cum-res-time-tab} and \Cref{tab:cl-cum-res-dy-tab} in \Cref{sec:app-sequence-mdoel}).
and the recommendation task below.

\subsection{Application 3: Large-Scale Recommendation}
\label{sec:exp-recommend}

Large-scale recommendation systems have seen quite a bit of change in categorical features. 
For example, new users or movies (categorical items) reach a streaming service, the recommender needs to incorporate them and make recommendations. We now demonstrate how PHE can assist recommendation systems in online learning.

\paragraph{Methods.} We treat the recommendation problem as a rating prediction problem, where the task is predicting the rating a user gives to a movie.
We combine PHE and Neural Collaborative Filtering \citep{he2017neural} as the backbone model.  
We assume all ratings are iid Gaussian distributed conditioned on user, movie, and movie-genre embeddings. And we model user and movie embeddings through PHE, denoted by $E_{\h_u}$ and $E_{\h_m}$, while movie-genres are encoded as multi-hot embeddings denoted by $\x$. Then the likelihood is $p_\theta(y|E_{\h_u}, E_{\h_m}, \x)$ with learnable parameters $\theta$. $\theta$ are the weights of a two-layer neural network. We model the mean of $y$ as the network output conditioned on input $[E_{\h_u}, E_{\h_m}, E_{\h_u} \odot  E_{\h_m}, \x]$.

\paragraph{Datasets.}
We apply the largest MovieLens-32m \citep{harper2015movielens} which contains 32 million ratings across over 87k movies and 200k users. These data were recorded between 1/9/1995 and 10/12/2023 for about 28 years.
Each piece of data is a tuple of \texttt{(userId, movieId, rating, timestamp)}, recording when and which rating a user gave a movie. Ratings range from 0 to 5 stars with half-star increments.

\paragraph{Experimental setups.}
In implementation, ratings are normalized to $[0,1]$ and are taken to be continuous albeit their increments are discrete. 
We simulate the experiment as in production -- online prediction along the timestamp. 
The model is pre-trained on the first five years of data and then perform predict-update online learning on a daily basis.
In this setup, both forgetting and adaptation in the hash embeddings are measured: the model should avoid forgetting for recurring users/movies and adapt for new users/movies.
Prediction error is evaluated by mean absolute error.

\paragraph{Results.}
The results of all compared methods are shown in \Cref{fig:all-tab-online} in supplement and the memory efficiency of PHE is reported in \Cref{tab:cum-res-dy-tab,tab:memory-cost}. 
It shows that PHE outperforms all deterministic hash embedding baselines (Fast/Medium/SlowAda) that have various forgetting-adaptation trade-offs. PHE also significantly outperforms the collision-free P-EE baseline. This is remarkable considering PHE consumes only 4\% of the memory of P-EE. 
EE, the deterministic version of P-EE, has worse performance, possibly due to overfitting.

\subsection{Additional Results}
\label{sec:exp-add}

We conducted additional experiments and presented the results in \Cref{sec:app-add-exp}. We showcased additional motivating examples beside \Cref{fig:group-update-perf}; demonstrated the \textbf{memory and hardware efficiency} of PHE in \Cref{tab:memory-cost}; analyzed \textbf{adaptation and forgetting} separately in \Cref{fig:streaming}; investigated classification and sequence modeling in classical \textbf{continual learning setup} in \Cref{fig:cum-res-time-tab}; performed \textbf{ablation studies} on the hash size $B$, the number of hash functions $K$, compared multiple \textbf{update schemes}, and the performance of \textbf{double-size} Ada baselines.%

\section{Conclusions}
\label{sec:concl}

In this work we unveiled the ineffectiveness of hash embeddings in online learning of categorical features. 
We proposed probabilistic hash embeddings (PHE) and addressed the problem of online learning. 
We showcased PHE is a plug-in module for multiple ML models in various domains and applications, allowing these models to learn categorical features in a streaming fashion.
We derive scalable inference algorithms to simultaneously learn the model parameters and infer the latent embeddings. 
Through Bayesian online learning, the model is able to adapt to new vocabularies without additional hyperparameters in a changing environment. We benchmark \method and baselines on large-scale public datasets to demonstrate the efficacy of our method. %

\bibliography{BIB/ref}

\newpage
\appendix
\onecolumn
\appendix

\section{Evidence Lower Bounds}
\label{app:elbo}

\subsection{Variational inference as KL divergence minimization}
\begin{align}
    &\min_q \KL(q_\lambda(E)|p(E|\gD))\\
    = &\min_q \E_{q_\lambda(E)}[\log q_\lambda(E) - \log p(E|\gD)] \\
    = &\min_q \E_{q_\lambda(E)}[\log q_\lambda(E) - \log p(E) -\log p(\gD|E) + \log p(\gD)]\\
    = &\max_q  \E_{q_\lambda(E)}[\log p(\gD|E)] - \KL(q_\lambda(E)|p(E)) \\
    = &\max_\lambda \gL(\lambda)
\end{align}

\subsection{Derivation of $\gL(\theta, \lambda, \phi)$ in \Cref{sec:exp-dy-temp-tab}}

\begin{figure*}
    \begin{center}
        \includegraphics[width=0.35\textwidth]{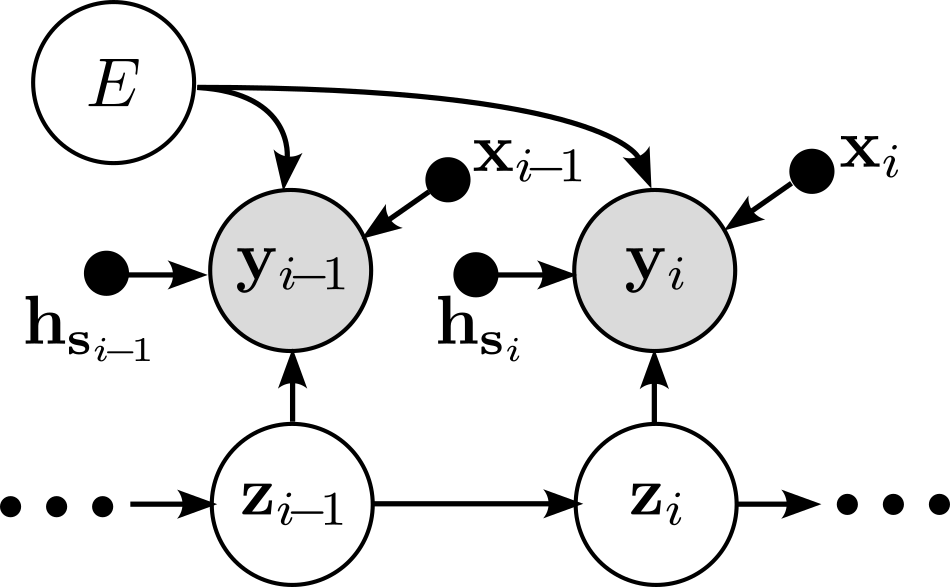}
        \vspace{-0.5em}
        \caption{
        A graphical model of a temporal sequence with PHE. The changing categorical values are contained in $\h_\s$.
        }
        \label{fig:seq-model-gen-proc}
    \end{center}
    \vspace{-1em}
\end{figure*}

The data generating process is summarized in \Cref{fig:seq-model-gen-proc}. We denote the model parameters relevant to the generating process by $\theta:=\{\theta_z, \theta_y\}$. 
To learn the model parameters, we maximize the marginal likelihood $p(\y_{\leq N}|\x_{\leq N}, \h_{\s_{\leq N}};\theta)$. 
Directly optimizing this marginal likelihood with the Expectation-Maximization (EM) algorithm is intractable. 
Therefore, we jointly learn the model parameters $\theta$ and infer the variational posteriors of latent variables $\{E, \z_{\leq N}\}$ using the variational EM algorithm. That is, we maximize the evidence lower bound (ELBO) $\gL(\theta, \lambda, \phi)$ with respect to model parameters $\theta$ and variational parameters $\{\lambda,\phi\}$. 

Denote all the history $\{\x_{\leq i}, \y_{\leq i}, E_{\h_{\s_{\leq i}}}\}$ until row $i$ by $O_i$. We find the optimal parameters by maximizing the marginal evidence $p(\y_{\leq N}|\x_{\leq N}, \h_{\s_{\leq N}};\theta)$. We take the logarithm of marginal evidence
\begin{align}
    &~\log p(\y_{\leq N}|\x_{\leq N}, \h_{\s_{\leq N}};\theta)\\
    =&~\log \int \frac{p(\y_{\leq N}, E|\x_{\leq N}, \h_{\s_{\leq N}};\theta)}{q_\lambda(E)}q_\lambda(E) dE \\
    \geq &~ \E_{q_\lambda(E)}[\log p(\y_{\leq N}, E|\x_{\leq N}, \h_{\s_{\leq N}};\theta) - \log q_\lambda(E)] \\
    = &~\E_{q_\lambda(E)}[\log p(\y_{\leq N}|\x_{\leq N}, E_{\h_{\s_{\leq N}}};\theta)] -\KL(q_\lambda(E)|p(E))\label{eq:app-elbo-0}
\end{align}
where the inequality follows from Jensen's inequality. Next, we apply the same trick for another time to find a lower bound of \Cref{eq:app-elbo-0}. Specifically, we will find a tractable lower bound to the conditional likelihood $\log p(\y_{\leq N}|\x_{\leq N}, E_{\h_{\s_{\leq N}}};\theta)$.

In the filtering setup, we note that $\log p(\y_{\leq N}|\x_{\leq N}, E_{\h_{\s_{\leq N}}};\theta)=\sum_{i=1}^N \log p(\y_i|\y_{<i}, \x_{\leq i}, E_{\h_{\s_{\leq i}}};\theta)$. If we can find a lower bound for each $\log p(\y_i|\y_{<i}, \x_{\leq i}, E_{\h_{\s_{\leq i}}};\theta)$, then the summation of the lower bounds is also a valid lower bound for $\log p(\y_{\leq N}|\x_{\leq N}, E_{\h_{\s_{\leq N}}};\theta)$.
\begin{align}
    &~\log p(\y_i|\y_{<i}, \x_{\leq i}, E_{\h_{\s_{\leq i}}};\theta)\\
    =&~\log \int \frac{p(\y_i, \z_i|\y_{<i}, \x_{\leq i}, E_{\h_{\s_{\leq i}}};\theta)}{q_\phi(\z_i|O_i)}q_\phi(\z_i|O_i) d\z_i \\
    \geq&~\E_{q_\phi(\z_i|O_i)}[\log p(\y_i|\y_{<i}, \x_{\leq i}, E_{\h_{\s_{\leq i}}}, \z_i;\theta)] - \KL(q_\phi(\z_i|O_i)|p(\z_i|O_{i-1}))\label{eq:app-elbo-i-1} \\
    \geq&~\E_{q_\phi(\z_i|O_i)}[\log p(\y_i|\y_{<i}, \x_{\leq i}, E_{\h_{\s_{\leq i}}}, \z_i;\theta)] - \E_{q(\z_{i-1}|O_{i-1})}\KL(q_\phi(\z_i|O_i)|p(\z_i|\z_{i-1};\theta_z))\label{eq:app-elbo-i}
\end{align}
\Cref{eq:app-elbo-i-1} to \Cref{eq:app-elbo-i} follows from the following inequality:
\begin{align}
    \KL(q_\phi(\z_i|O_i)|p(\z_i|O_{i-1})) \leq \E_{q(\z_{i-1}|O_{i-1})}\KL(q_\phi(\z_i|O_i)|p(\z_i|\z_{i-1};\theta_z))
\end{align}
because
\begin{align}
    &~\KL(q_\phi(\z_i|O_i)|p(\z_i|O_{i-1})) \\
    =&~\E_{q_\phi(\z_i|O_i)}[\log q_\phi(\z_i|O_i) - \log p(\z_i|O_{i-1})] \\
    =&~\E_{q_\phi(\z_i|O_i)}\left[\log q_\phi(\z_i|O_i) - \log \E_{q(\z_{i-1}|O_{i-1})}[p(\z_i|\z_{i-1};\theta_z)]\right]\label{eq:kalman-pred}\\
    \leq &~\E_{q_\phi(\z_i|O_i)}\left[\log q_\phi(\z_i|O_i) - \E_{q(\z_{i-1}|O_{i-1})}[\log p(\z_i|\z_{i-1};\theta_z)]\right] \\
    =&~\E_{q(\z_{i-1}|O_{i-1})q_\phi(\z_i|O_i)}[\log q_\phi(\z_i|O_i) - \log p(\z_i|\z_{i-1};\theta_z)]\\
    =&~\E_{q(\z_{i-1}|O_{i-1})}\KL(q_\phi(\z_i|O_i)|p(\z_i|\z_{i-1};\theta_z)).
\end{align}

Then \Cref{eq:app-elbo-i} is the conditional ELBO $\gL_i(\theta,\phi|E)$ in \Cref{sec:exp-dy-temp-tab}. Plug \Cref{eq:app-elbo-i} in \Cref{eq:app-elbo-0}, we have
\begin{align}
    &~\E_{q_\lambda(E)}[\log p(\y_{\leq N}|\x_{\leq N}, E_{\h_{\s_{\leq N}}};\theta)] -\KL(q_\lambda(E)|p(E)) \\
    \geq &~ \E_{q_\lambda(E)}\left[\sum_{i=1}^N \gL_i(\theta,\phi|E)\right] -\KL(q_\lambda(E)|p(E))
\end{align}
which is our objective function $\gL(\theta,\phi,\lambda)$ in \Cref{sec:exp-dy-temp-tab}.

\subsection{$\gL(\theta, \phi, \lambda)$ as a Variational EM Algorithm}
\label{app:var-em-proof}

\textit{Why is maximizing $\gL(\theta, \phi, \lambda)$ a meaningful objective as a variational expectation-maximization algorithm?}
We start with a general latent variable model $p_\theta(x,z)=p(z)p_\theta(x|z)$ and infer the posterior $p_\theta(z|x)$.
\begin{align*}
    &~\KL(q_\lambda(z)|p_\theta(z|x)) \\
    :=&~\E_{q_\lambda(z)}[\log q_\lambda(z) - \log p_\theta(z|x)] \\
    = &~\E_{q_\lambda(z)}[\log q_\lambda(z) - \log p_\theta(x,z) + \log p_\theta(x)] \\
    = &~ -\gL(\lambda, \theta) + \log p_\theta(x)
\end{align*}
Re-ordering the equation yields
\begin{align*}
    \gL(\lambda, \theta) = \log p_\theta(x) -\KL(q_\lambda(z)|p_\theta(z|x)),
\end{align*}
which shows that maximizing the ELBO $\gL(\lambda, \theta)$ is equivalent to both maximizing the marginal likelihood $p_\theta(x)$ and minimizing the inference gap $\KL(q_\lambda(z)|p_\theta(z|x))$.

Then, with the same procedure as above, two facts follow: 1) maximizing $\gL_i(\theta,\phi|E)$ is equivalent to maximizing the conditional likelihood $\log p(\y_i|\y_{<i},\x_{\leq i}, E_{\h_{\s_{\leq i}}};\theta)$ and minimizing the inference gap $\KL(q_\phi(\z_i|O_i)|p(\z_{i}|O_{i};\theta))$ simultaneously; 2) maximizing \Cref{eq:app-elbo-0} is equivalent to maximizing $\log p(\y_{\leq N}|\x_{\leq N},\h_{\s_{\leq N}};\theta)$ and minimizing the inference gap $\KL(q_\lambda(E)|p(E|\y_{\leq N},\x_{\leq N},\h_{\s_{\leq N}};\theta))$ simultaneously. Since maximizing $\gL(\theta, \phi, \lambda)$ optimizes both $\gL_i(\theta,\phi|E)$ and \Cref{eq:app-elbo-0}, we conclude our objective function will optimize all the mentioned aspects above.

\subsection{Derivation of $\gL^{(1)}(\lambda;\theta^*,\lambda_0^*, \phi^*)$  in \Cref{sec:exp-dy-temp-tab}}
We only adapt the probabilistic hash embedding $E$. Similar to Bayesian online learning where the previous posterior is used as the new prior, we use the previous approximate posterior $q_{\lambda^*_0}(E)$ as the new prior for dataset $\gD_1$ and fix all the other model parameters $\theta^*,\phi^*$. The derivation is the same as the one for \Cref{eq:app-elbo-0} except we replace $p(E)$ with $q_{\lambda_0^*}(E)$. We only update $\lambda$ to acquire the new posterior in the optimization.

\section{Proof of \Cref{prop:bayes-online-learning}}
\label{app:proof-prop3.1}
\begin{proof}
The proof is simple and based on repetitive applications of Bayes rule. 
Let $\gD_{\pi_i}$ denote the subset of data arrived before (and includes) $\pi_i$. We assume the data arrive one by one and we cannot store previous data.
\begin{align*}
    p(E|\gD_{\pi_N}) &\propto p(E|\gD_{\pi_{N-1}})p(y_{\pi_N}|E,s_{\pi_N}) \\ 
    &\propto p(E|\gD_{\pi_{N-2}})p(y_{\pi_N}|E,s_{\pi_N})p(y_{\pi_{N-1}}|E,s_{\pi_{N-1}}) \\
    & ...\\
    &\propto p(E)\prod_{i=1}^N p(y_{\pi_i}|E,s_{\pi_i})
\end{align*}
Because $\prod_{i=1}^N p(y_{\pi_i}|E,s_{\pi_i})=\prod_{i=1}^N p(y_i|E,s_i)$, we conclude $p(E|\gD_{\pi_N})$ is the same as $p_\text{batch}(E|\gD)$.
\end{proof}

\section{Limitation}
\label{app:limitation}
We discuss two limitations in this study.
1) The need to pre-specify embedding parameters $K$ and $B$ can limit model adaptability if actual category counts exceed expectations (though this can be mitigated through retraining or adding a forgetting mechanism to KL term in \cref{eq:elbo-online-sec3.3}; 2) Restricting updates to embeddings $E$ while fixing network parameters $\theta$ may speed up the consumption of $E$ under strong distribution shifts, blocking new learnings and forgetting old. Meanwhile, Our ablation in \cref{tab:ablation-incre-only-update} suggests that updating $\theta$ requires more sophisticated strategies to prevent forgetting.

\section{Related work}
\label{app:related}
\begin{table}[htbp]
\begin{center}
\begin{tabular}{ l| c | c | c | c | c}
        & $D0$ & $D1$ & $D2$ & $D3$ & $D4$ \\ 
\hline  
        Changing vocabulary &  &  & \CheckmarkBold & \CheckmarkBold & \CheckmarkBold\\
        Timestamped       &  & \CheckmarkBold & & \CheckmarkBold & \CheckmarkBold\\
        Multi-task        &  &  &  &  & \CheckmarkBold\\ \hline
\end{tabular} 
\caption{Tabular datasets can be categorized into five categories ($D0-D4$) based on combinations of three characteristics, i.e., whether their categorical feature vocabulary dynamically expands over time, whether they contain a specific timestamp column, and whether their nature is multi-task. For example, datasets without all these characteristics are considered static ($D0$). While existing works mainly consider $D0$ and $D1$, \method fits all dataset types ($D0-D4$) and specifically highlights the unique applicability for dynamic and temporal tabular data types ($D1-D4$).
}
\label{tab:temp-tab-data-type}
\end{center}
\end{table}

We extend the discussion in \Cref{sec:background} and survey more related works. In a nutshell, our \method applies to all tabular data types in \Cref{tab:temp-tab-data-type} (i.e., $D0-D4$) while existing works are targeted to $D0$ or $D1$. 

Our work deals with multi-task dynamic temporal tabular data. Our method has two major components: the probabilistic hash embeddings that learn categorical feature representations and the latent variable model for multi-task temporal tabular data. Next, we discuss the main related works.

\paragraph{Hash features.} PHE is motivated by hashing tricks. \citet{weinberger2009feature} proposed to use one hash function to map categorical features to a one-hot hash embedding of length $B$, which is the bucket size. The drawback is the embedding size is too large because there is only one hash function and that requires a large bucket size $B$ to get rid of collision. Bloom Embeddings~\citep{serra2017getting} is based on Bloom filters and achieves efficient computation while maintaining a compact model size. Other previous work on using hashing tricks to generate features focuses on using a smaller number of embedding-related parameters to achieve the same performance as using one-hot encoding.  Hash embeddings or unified embeddings~\citep{tito2017hash,cheng2023efficient} use a shared embedding table for all categorical features and multiple hashing functions as indices of the embedding table, reducing the possibility of collision. Hash embeddings are designed for stationary vocabularies, emphasizing small parameter sizes. We generalize hash embeddings to a probabilistic version that enables us to learn changing vocabularies via Bayesian online learning. Composition Embeddings~\citep{shi2020compositional} use multiple hash embedding tables; in contrast, PHE uses one shared embedding table, further reducing the memory cost. Wolpertinger~\citep{dulac2015deep} and Deep Hash embedding~\citep{kang2021learning} use a deep neural network to encode features into real-valued embeddings. In a changing vocabulary setup, the drawback is the need to modify the whole neural network to incorporate new string features, even though there is only one new feature.  Different from previous works, our method emphasizes the usage of hash embeddings in dynamic tabular data with changing vocabularies. In the meantime, the model architecture remains stable, and only partial parameter updates are required.

\paragraph{Generative models for tabular data.} Recent research on generative models of non-temporal tabular data focuses on modeling multi-modality or heterogeneity but overlooks the sustainable representations for dynamically expanded vocabularies. These works rely on one-hot encoding for categorical features. \citet{xu2019modeling} learns VAE and GAN-based tabular data generator while conditioning on discrete categorical features. Later works rely on GAN to design tabular data generators~\citep{liu2023tabular,zhao2021ctab}. \citet{kotelnikov2023tabddpm} extend diffusion models to tabular data.

\paragraph{Temporal tabular data models.} To our knowledge, there isn't a sequence model designed for multi-task temporal tabular data, although some previous works have the potential to extend to tabular data. \method extends Deep Kalman Filters~\citep{krishnan2015deep} to be applicable for multi-task, temporal, and dynamic tabular data, while the original Deep Kalman Filters do not explicitly consider the multi-task and dynamic vocabulary property of the tabular data. \citet{girin2021dynamical} survey a list of latent variable sequence models that are possible to be extended to tabular data, although most of them are designed for speech or video data.

\paragraph{Others.} The setup of learning dynamic tabular data with changing vocabularies shares the similarity to continual learning and Bayesian online learning~\citep{kirkpatrick2017overcoming, wang2023comprehensive, zenke2017continual, nguyen2018variational, li2021detecting}, but the difference is our formulation is a novel dictionary- or vocabulary-incremental setup for tabular data. Besides, \citet{kireev2023transferable} learn transferable robust embeddings for categorical features.  \citet{yin2020tabert,iida2021tabbie} design objective functions for representation learning on tabular data using large-language models. \citet{arik2021tabnet} and \citet{huang2020tabtransformer} use the one-hot encoder to learn categorical feature embeddings before input to a transformer module.

\paragraph{Discussions on alternative designs and shortcomings.} We acknowledge that alternative solutions may exist, e.g., encoding string features with a character-level recurrent neural network or using a popularity-based token-level one-hot encoder. In our considered aspects, for example, long-tailed data distributions are commonly seen in applications, probabilistic hash embedding stands out with simplility and continual learning capability. Hash features~\citep{weinberger2009feature,cheng2023efficient} is memory inefficient. Incremental one-hot embeddings are also inefficient for dynamic tabular data, because the model parameters expand unbounded, resulting in storage inefficient and slow computation. Deep hash embedding~\citep{kang2021learning} and other methods in the same fashion are computationally inefficient. One needs to adapt the whole neural network even when adding one new category. In contrast, one only needs to adapt the corresponding embeddings in probabilistic hash embedding. 

\paragraph{Handling hashing value collisions.} Collision of hash values could happen among popular, important categories. To address this issue, we can select the desired hash functions that avoid important collisions before applying the hash functions. In addition, users come and go fast, and collisions may become unimportant over time.

\section{An example tabular data of changing categorical features}
\label{sec:app-ttd-snippet}
\begin{table*}[t!]
\caption{A tabular data snippet from the Retail dataset. The columns are either categorical, numeric, or timestamp. The rows corresponds to sale records.
\texttt{StockCode} stores product ID. \texttt{Quantity} stores the sales. ``?'' denotes missing values.
The task is to predict the sales for each product.
}
\label{tab:retail-snippet}
\centering
\small
\begin{tabular}{llllll}
\toprule 
StockCode & Date & UnitPrice & CustomerID & Country & Quantity \\
\midrule
85123A &2010-12-01 08:26:00 &2.55 &17850 &United Kingdom &6\\
84406B &2010-12-01 08:26:00 &2.75 &17850 &United Kingdom &6\\
21724 &2010-12-01 08:45:00 &0.85 &12583 &France &12\\
21791 &2010-12-01 10:03:00 &1.25 &12431 &Australia &12\\
22139 &2010-12-01 11:52:00 &0.55 &? &United Kingdom &56\\
\bottomrule
\end{tabular}
\end{table*}

We will explain the concepts related to this work through an example tabular data snippet (\Cref{tab:retail-snippet}). Tabular data contains two dimensions--rows and columns. Any stored information can be located by specifying the row and column indices. We can classify columns into three types: \emph{categorical}, \emph{numeric}, and \emph{timestamp}. 
A categorical column represents a discrete nominal feature, usually recorded in text strings and therefore hashable; A numeric column corresponds to a numeric feature, usually represented by float or integer values; and a timestamp column records the timestamp when a row is created. For instance, in \Cref{tab:retail-snippet}, there are six columns, among which \texttt{StockCode}, \texttt{CustomerID}, and \texttt{Country} are categorical columns, \texttt{UnitPrice} and \texttt{Quantity} are numeric columns, and \texttt{Date} is a timestamp column. 
Some columns are of particular interest and one may want to predict those based on others. We refer to those columns as \emph{predicted columns}. Predicted columns can be either categorical or numeric, depending on task requirements.
Rows with similar timestamps usually exhibit correlations. But these correlations may change over time.

Some tabular data is multi-task-oriented. For example, in \Cref{tab:retail-snippet}, one may be interested in predicting future selling quantity based on historical transactions for each product. In this case, different product IDs in \texttt{StockCode} suggest different tasks. We refer to the categorical columns consisting of task identifiers as \emph{global columns} and other categorical columns as \emph{local columns}. We express this type of tabular data \emph{multi-task}. Each task may have specific column relationships.

All unique items in a categorical column constitute its \emph{vocabulary}. When new items join into the column, we say it has a \emph{changing} or \emph{dynamic vocabulary}. 
\footnote{
  We assume the tabular structure is fixed, i.e., the number of columns, column names, and types are fixed. We also assume categorical features are single-valued. But our work is compatible with multi-valued features.
}

{\color{black}

\section{A simple example to understand why PHE is superior to DHE}
\label{sec:theory}

\begin{figure*}[t]
    \begin{center}
           \includegraphics[width=0.7\textwidth]{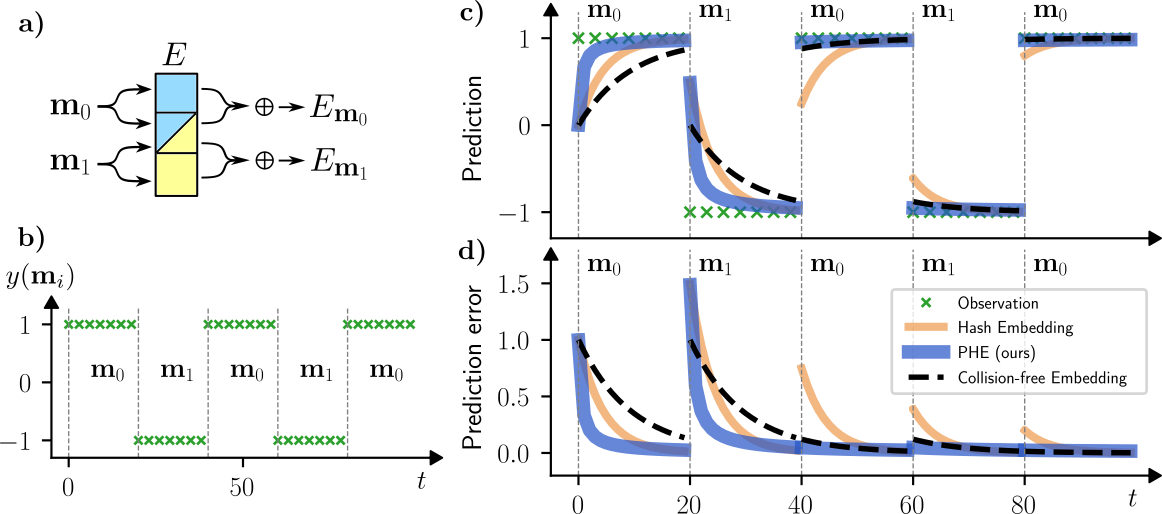}
        \vspace{-0.5em}
        \caption{ Forgetting in online learning using deterministic hash embedding on synthetic data. (The complete setting is described in \Cref{sec:minor_theory}.)
        The task is predicting a scalar (regression problem) with the covariate being a categorical variable that takes one of two values of $\m_0$ or $\m_1$.
        \textbf{a)} shows the embedding matrix $E$ of size $3 \times 1$. Here the number of buckets $B=3$ and $d=1$. The two hash function maps $\m_0$ to $0$ and $1$ respectively and maps $\m_1$ to $1$ and $2$ respectively.
         \textbf{b)} shows the online samples where the covariate alternates between $\m_0$ and $\m_1$ and the corresponding target $y(\m_i)$ takes values in $1$ and $-1$. 
         \textbf{c)} shows the prediction of a probabilistic hash embedding table (blue) trained using Bayesian online learning and a deterministic hash embedding (DHE) table (yellow) trained using online gradient descent. \textbf{d)} plots the prediction error. From these figures we observe that PHE's prediction error converges to $0$ much quicker than DHE. After every $20$ samples when the covariate changes, there is a big jump in DHE error, exhibiting forgetting while the PHE has no error spikes after it has encountered both the categorical values.
       }
           \label{fig:synthetic-data-forget} 
    \end{center}
    \vspace{-0.5em}
\end{figure*}

In this section, we consider a simple linear Gaussian model that we can analyze in closed form to illustrate why having deterministic hash embeddings that are updated in an online fashion is prone to \emph{forgetting}. The crux of our calculations is the fact that distinct categorical items share representations due to partial hash collisions. Thus, when trained online, the shared features shows a bias to work well for the categorical item that was most recently seen, rather than be optimized for the overall data distribution seen so far, leading to the forgetting behaviour. However, we will show that Bayesian hash embedding does not suffer this, because it is well known that if exact online posterior can be computed (which in our linear Gaussian setup is easy to do), the online posterior is identical to the offline one.\footnote{
  Note that this section has a slightly different notation from the main text, but the content is self-contained. Readers can also match the notation by noting $X:=\m$ and the input variable value has $0:=\m_0$ and $1:=\m_1$.
}

\textbf{A simple linear-Gaussian model}

Consider a simple situation of regression with input variable $X \in \{0,1\}$ taking one of two categorical values and the target $Y \in \mathbb{R}$ is real-valued. The conditional distribution of $Y$ is a gaussian distribution with the mean being $1$ when $X=0$ and mean being $-1$ when $X=1$. We further assume that the variance of $Y$ is $\sigma^2 \approx 0$ is tiny, In notation terms, the true distribution of  $Y \vert X=0 \sim \mathcal{N}(1,\sigma^2)$, while the distribution of $Y \vert X=1 \sim \mathcal{N}(-1,\sigma^2)$, where $\sigma$ is a fixed and small. We do not specify the distribution of the covariate $X$ just yet and defer that to the sequel.

\textbf{The predictive model based on hash embedding}

Given labeled data $(X,Y)$, we aim to learn a predictor $f(X)$ that predicts $Y$ given $X$. To build the predictor we use a simple hash embedding model. Specifically, we assume that the predictor $f(\cdot)$ is parameterized by a $3 \times 1$ embedding matrix $E$. Although technically this is a vector, we still denote it as an `embedding matrix' to be consistent with the rest of the exposition. Denote by $e^{(0)}, e^{(1)}$ and $e^{(2)}$ as the three rows of this matrix which are the `embedding vectors' of the three hash values. Thus, in the notation of our model, this embedding matrix is made of $B=3$ buckets with the dimension $d=1$. The model $f(\cdot)$ uses two hash functions $h_i(\cdot) : \{0,1\} \to \{0,1,2,\}$ to map the categorical variable $X$ into a hash value. Without loss of generality, we assume that $h_1(0) = 0, h_2(0) = 1, h_1(1) = 1, h_2(1) = 2$. Given this, the predictive model $f(X) := e^{(h_1(X))} + e^{(h_2(X))}$ is a simple linear sum of the two hash embedding of the input based on the two hash functions $h_1(\cdot)$ and $h_2(\cdot)$. This is a simple example of the general class of models where the predictor $Y$ is a linear function of the embedding vectors of the categorical input $X$ computed using the different hash functions. Although simple, this example illustrates the phenomenon that emerges of learning categorical variables in an online fashion since the embedding vector $e^{(1)}$ influences both $X=0$ through hash function $h_1(\cdot)$ and $X=1$ through hash function $h_2(\cdot)$.

\textbf{An online interaction setting}

We consider the following online prediction protocol. At each time $t = 1,2, \cdots$, the environment samples $X_t$ from a distribution over $\{0,1\}$ and produces to the predictor. The predictor then predicts $\widehat{Y}_t := f_{t}(X_t)$ and is then shown the true label $Y \in \mathbb{R}$. The predictor incurs loss $l_t := \frac{1}{2}(Y_t - \widehat{Y}_t)^2$ and uses the observed $Y_t$ to update the predictor to $f_{t+1}(\cdot)$. 

The only learnable parameters of the predictor is the embedding matrix $E$. Thus the predictor at time $t$ denoted by $f_{t}(\cdot)$ is parametrized by the state of the embedding matrix $E_t$ with its three rows denoted by $e^{(i)}_t$ for $i \in \{0,1,2\}$. 

\textbf{Update the hash embedding matrix through Online Gradient Descent (OGD)}

In order to demonstrate that the hash embeddings can lead to forgetting, we will assume that they are updated through standard online gradient descent. Observe that at time $t$, if $X_t = 0$, then $\widehat{Y}_t = e^{(0)}_t + e^{(1)}_t$. The instantaneous loss at time $t$ is given by $l_t = \frac{1}{2}(\widehat{Y}_t - Y_t)^2$. Thus, the gradients $\frac{\partial l_t}{\partial e^{(0)}} = \frac{\partial l_t}{\partial e^{(1)}} = (e^{(0)}_t + e^{(1)}_t - Y_t) $, if $X_t=0$. Thus, assuming that the embedding matrix $E_t$ is updated online using OGD at a fixed learning rate $\eta \in \mathbb{R}$ leads to the following update equations
\begin{equation*}
e_{t+1}^{(0)} = \begin{cases}
 e_{t}^{(0)} - \eta ((e^{(0)}_t + e^{(1)}_t - Y_t) ),& X_t = 0\\
e_{t}^{(0)}, &X_t = 1.
\end{cases}
\end{equation*}

Similarly the update equations for the other two embedding vectors are as follows. 

\begin{equation*}
e_{t+1}^{(1)} = \begin{cases}
 e_{t}^{(1)} - \eta ((e^{(0)}_t + e^{(1)}_t - Y_t) )& X_t = 0\\
e_{t}^{(1)} - \eta ((e^{(1)}_t + e^{(2)}_t - Y_t) ) &X_t = 1
\end{cases}
\end{equation*}

\begin{equation*}
e_{t+1}^{(2)} = \begin{cases}
 e_{t}^{(2)} & X_t = 0\\
e_{t}^{(2)} - \eta ((e^{(1)}_t + e^{(2)}_t - Y_t) ) &X_t = 1
\end{cases}
\end{equation*}

These update equations for the embedding shows that $e^{(1)}$ which is shared for both $X=0$ and $X=1$ gets updated all the time, while $e^{(0)}$ is only updated if $X=0$ and similarly $e^{(1)}$ is only updated if $X=1$. 

\textbf{A non-stationary distribution for the co-variates $X$}

Consider a setting where the first $N$ inputs consists of $X_t = 0$ for all $t \in \{1,\cdots,N\}$, followed by another $N$ inputs consisting of $X_t=1$ for all $t \in \{N+1, \cdots, 2N\}$. In these discussions we will assume $N$ is large enough and the learning rate $\eta$ is appropriately tuned to make the variance of the predictor to be small.
If all the $2N$ samples were shown to a training algorithm, it could have (near) perfectly estimated the embedding matrix $\widehat{E}$, i.e., for a $X$ that is sampled from $\{0,1\}$ that is equally likely (matching the training data distribution of equal number of $0$ and $1$), the expected excess loss will be arbitrarily small (assuming $N$ is sufficiently large). We will show in the calculations below that if instead the embedding matrix was learnt using OGD, even if $N$ is large enough, the learnt model at the end will have a constant excess risk when the test input $X$ is sampled with equal probability among $\{0,1\}$.

\textbf{Analyzing the OGD update equations}

To see this, we make some simplifying assumptions. First is that $\sigma = 0$, i.e., conditioned on $X$, $Y$ is deterministic.  Second is a symmetric starting point of $e_0^{(i)} = 0$ for all $i \in \{0,1,2\}$. It is easy to observe that both of these assumptions do not change the the observation we will make, but makes the exposition easier. Thus, at the end of the first $N$ samples, we will have $e_{N+1}^{(2)} = 0$ and $e_{N+1}^{(0)} = e_{N+1}^{(1)} \approx 1/2$. This follows as $N$ is large and the noise $\sigma$ is $0$, thus leading OGD to converge to a local minima of the loss function. Any embedding matrix with $e^{(0)} + e^{(1)} = 1$ is a local-minimum of the loss function and thus at the end of time $N+1$, OGD will result in $e_{N+1}^{(0)} + e_{N+1}^{(1)} \approx 1$. Since the initialization and the loss function is symmetric in the arguements $e_{t}^{(0)}=e_{t}^{(1)}$ will hold for all $t \leq N$. 

At time $t=N+1$, the $N$ observed samples corresponds to $X=0$. Thus, the prediction error for $X=0$ by this learnt model $\widehat{f}_{N+1}(X)$ is small, i.e., the excess risk $(f_{N+1}(X) - 1)^2 \approx 0$.

Now consider the times $t=N+1$ till $t=2N$. During this period, the gradients will not impact $e^{(0)}$, i.e., $e_{N+1}^{(0)} = e_{2N+1}^{(0)} \approx 1/2$. However, $e^{(2)}$ and $e^{(3)}$ are no longer symmetric. But one can work out the recursion for their evolution since the gradients are the same. 

In particular, for any time $t \in \{N+1, \cdots, 2N\}$, the observed $X_t = 1$. Thus, the gradient of $e^{(1)}$ and $e^{(2)}$ at all times $t \in \{N+1, \cdots, 2N\}$ is the equal to $(e^{(1)}_t + e^{(2)}_t + 1)$. 
Thus, under the OGD update equations, for all times $t \in \{N+1, \cdots, 2N\}$, the equality $e^{(1)}_{t+1} - e^{(2)}_{t+1} = e^{(1)}_{t} - e^{(2)}_{t}$, holds. 
Since at time $N+1$, we have $e^{(1)}_{N+1} \approx 1/2$ and $e^{(2)}_{N+1} = 0$, we have that $e^{(1)}_{2N+1} - e^{(2)}_{2N+1} \approx 1/2$.  On the other hand, if $N$ is large, we know that OGD will converge to a local minima, i.e., $e^{(1)}_{2N+1} + e^{(2)}_{2N+1} \approx -1$. These two equations in the variables $e^{(1)}_{2N+1} , e^{(2)}_{2N+1}$ gives  $e^{(1)}_{2N+1} \approx -1/4$ and $e^{(2)}_{2N+1} \approx -3/4.$

\textbf{Concluding that the updates leads to forgetting the representation for $X=0$}

Thus at the end at time $2N+1$, after having seen the first $N$ samples of $X=0$ and the last $N$ samples of $X=1$, the predictor is such that $\widehat{f}_{2N+1}(0) \approx 1/4$ and $\widehat{f}_{2N+1}(1) \approx -1$. 
However, note that the true label when $X=0$ is $1$ while when $X=1$ is $-1$. Thus, the predictor $\widehat{f}_{2N+1}(\cdot)$ has near zero prediction error when $X=1$. However, when $X=0$, the loss given by $(\widehat{f}_{2N+1}(0) - Y)^2 \approx (1/4 - 1)^2 \approx 9/16$ is a constant. 

This shows the discrepancy between a model trained offline using all the $2N$ samples and the model trained online where the first $N$ samples all correspond to $X=0$ and the last $N$ samples correspond to $X=1$. The offline model will converge to a local minima in which the prediction error for both $X=0$ and $X=1$ will be small, while the online model converges to a solution where the prediction error for the categorical variable that was not seen recently is high.

\textbf{Arguing that online Bayesian model does not lead to forgetting}

A Bayesian method to `learn' the embedding matrix is to posit a prior distribution $p(E)$ for the emebedding matrix and then given the data $\mathbf{X}$ compute the posterior distribution $p(E \vert \mathbf{X})$. We will say that the Bayesian learning does not forget, if the posterior distribution computed based on all the $2N$ samples $(X_1, Y_1), \cdots, (X_{2N}, Y_{2N})$ shown up-front matches the posterior distribution computed in an online fashion. However, from classical results in online Bayesian learning, it is well known that if one can compute the exact posterior $p(E \vert X_1, \cdots, X_t)$ at all times $t$, then the posterior  at time $2N$ is identical to the one that an offline algorithm would have computed had it seen all the $2N$ samples at once. Thus, if the exact posterior can be computed at each time, then there is no forgetting in the Bayesian mechanism.

Thus in this section, we showed through a simple linear-gaussian model, that online updating of hash embedding matrix leads to forgetting while a bayesian updating of the embedding matrix does not lead to forgetting. In order to demonstrate this, we defined forgetting to not occur if the model learnt at the end of seeing each online sample one by one is close to the model learnt had all the samples been available up-front. Further, we show in experiments that this insight holds even in more complex scenarios where exact Bayesian posterior cannot be computed, but only an approximation through variational inference can be done.

}

\section{Experimental Details}
\label{app:exp}

\subsection{An efficient embedding fetch schemes}
\label{app:trade-off}
When implementing the hash embedding fetching module, there are two available schemes: scheme one is first to sample a whole hash table $E$ and then fetch the corresponding embeddings $E_{\h_\s}$ (as \Cref{eq:scheme-1}); scheme two is first to fetch the distribution $p(E_{\h_\s})$ and then sample $E_{\h_\s}$ (as \Cref{eq:scheme-2}).
\begin{align}
    p(\x|\s) = p(\x|\h_\s) &~= \E_{p(E)}[p(\x|E,\h_\s)]\approx p(\x|E,\h_\s) \label{eq:scheme-1}\\
        &~= \E_{p(E_{\h_\s})}[p(\x|E_{\h_\s})]\approx p(\x|E_{\h_\s}) \label{eq:scheme-2}
\end{align}

The two schemes lead to the same results, but scheme two is more memory-efficient as it does not need to sample the whole embedding table. Thus in practice, we apply \Cref{eq:scheme-2}.

\subsection{Hardware Information}
We train and test our model on GPUs (RTX 5000) and use the deep learning framework PyTorch to enable efficient stochastic backpropagation. In all supervised learning experiments, the total elapsed wall time (training and testing) for \method is less than half an hour, and the finetune baseline runs slightly faster. In the sequence modeling experiments, \method runs about one hour since Retail is a large dataset and has over 500k records. In the recommendation experiments, it takes about two hours for all methods.

\subsection{Details for Classification Experiments}
\label{sec:app-classification}

The four public datasets all can be found online: Adult\footnote{
  \url{https://archive.ics.uci.edu/dataset/2/adult}
}, Bank\footnote{
  \url{https://archive.ics.uci.edu/dataset/222/bank+marketing}
}, Mushroom\footnote{
  \url{https://archive.ics.uci.edu/dataset/73/mushroom} We also follow the recommendation and only use \texttt{odor} as the feature.
}, and Covertype\footnote{
  \url{https://archive.ics.uci.edu/dataset/31/covertype}
}. 
Specifically, Adult has 14 columns and 48,842 rows containing demographic information. 
The task is to predict whether or not a person makes over \$50K a year; Bank has 16 columns and 45,211 rows to predict if a client will subscribe to a term deposit; In Mushroom, of 22 discrete columns and 8,124 rows, the goal is to predict whether a mushroom is poisonous; Covertype, involving 12 columns and 581,012 rows, is to predict which forest cover type a pixel in a satellite image belongs to. 

Regarding model architecture, we concatenate all category embeddings as well as continuous features as input to a deterministic one-layer neural network, followed by a softmax activation function.
For PHE and P-EE, we stress that only embeddings are probabilistic and neural network weights are deterministic. 
We use negative cross entropy as the objective function assuming the targets follow categorical distributions.

We apply the following criterion when selecting a categorical column to have a dynamic vocabulary. We select the column to be dynamic if the weights of the column features have large scales when fitting a logistic regression model on the outputs. Specifically, we first use one-hot encodings to represent categorical items, and then fit a logistic regression model on the targeted outcomes. Finally, we select a column to be incremental if its corresponding categorical features have large weights because the weights in linear regression models can be interpreted as feature importance. Following this procedure, we select \texttt{education}, \texttt{poutcome}, \texttt{odor}, and \texttt{wilderness} column for the four datasets respectively. See detailed group information in \Cref{fig:group-info}.

For the continual learning setup in \Cref{sec:app-continual-learn}, 
we first randomly and evenly split the categorical features of the selected column into disjoint groups, then partition the original dataset according to the groups. Based on the column dictionary size, we split Adult/Bank/Mushroom/Covertype into five/four/four/four disjoint groups. We randomly split each group into training and testing subsets where the training subset takes two-thirds of the total data and the testing subset takes the remaining one-third. 
We sequentially fit the prediction model to each non-overlapped group. 
The goal is to have high accuracy for all groups after sequential updates. 
Therefore, after fitting the model on the current group's training data, we report the average accuracy on all previous groups' test data. 

\paragraph{Evidence lower bound.} We first present the objective function of the latent variable supervised learning model. %
Similarly to \Cref{eq:app-elbo-0}, we can derive the objective function as the evidence lower bound of $\sum_{i=1}^N\log p(\y_i|\x_i, \h_{\s_i};\theta)$:
\begin{align}
    \gL(\theta,\lambda) = \E_{q_{\lambda}(E)}\left[\sum_{i=1}^N \log p(\y_i|\x_i,E_{\h_{\s_i}};\theta)\right] - \KL(q_\lambda(E)|p(E))\label{eq:dytab-elbo}
\end{align}

For online adaptation to dataset $\gD_1$ of size $N_1$, we fix the classifier parameters and only adapt the hash embedding table $E$. Denote the pre-trained parameters by $\theta^*$ and $\lambda_0^*$. Treat the previous posterior $q_{\lambda_0^*}(E)$ as the current prior, we can write down the objective function
\begin{align}
    \gL^{(1)}(\lambda;\theta^*,\lambda_0^*)=\E_{q_{\lambda}(E)}\left[\sum_{i=1}^{N_1} \log p(\y_i|\x_i,E_{\h_{\s_i}};\theta^*)\right] - \KL(q_\lambda(E)|q_{\lambda_0^*}(E))\label{eq:dytab-elbo-adapt}
\end{align}

\paragraph{Implementation details and hyperparameters.} We implement the aggregation function $g$ as a weighted sum where the weights are parameters of $g$. Specifically, we have another random table $W\in\sR^{P\times K}$ whose distribution is $p(W)$ and a hash function $h^{(W)}:\gS\rightarrow \sN_{<P}$ such that $h^{(W)}_\s$ indexes the rows of $W$, noted by $W_{h^{(W)}_\s}\in\sR^K$. $W_{h^{(W)}_\s}$ serves as the weights for the $K$ hash embeddings (see \Cref{fig:hash-emb}). Then $g(E_{h_\s^{(1)}}, \dots, E_{h_\s^{(K)}})=\sum_{k=1}^K W_{h^{(W)}_\s}^kE_{h_\s^{(k)}}$ where $W_{h^{(W)}_\s}^k$ is the $k$th value of vector $W_{h^{(W)}_\s}$. During inference, we infer the posteriors of both $E$ and $W$. 

For all tabular datasets except Mushroom, we set $B=7,K=3,d=20,P=11$ (whose supported dictionary size is $P\times B^K=3773$, which is ten times larger than the vocabulary size of the Adult dataset). 
We tried these values on Adult when setting the group size to be one (i.e., the static supervised learning setup) and found the resulting accuracy (about $84\%$) is comparable to the public results on this dataset\footnote{
  See the baseline model performance in \url{https://archive.ics.uci.edu/dataset/2/adult}
}. We then use this same parameter setup on all other tabular data supervised learning experiments. 
For Mushroom, we use a much smaller model size and set $B=5,K=3,d=5,P=1$, because only one feature is used in the experiment.

\paragraph{Optimization.} We use Adam stochastic optimization with a learning rate of 0.01 and a minibatch size of 128 in all experiments for both our method and baselines. For other hyperparameters of Adam, we apply the default values recommended in the PyTorch framework. When selecting these values, we fixed the minibatch size 128 and searched the learning rate (0.001, 0.005, 0.01, 0.05, 0.1) on the Adult dataset. We found the learning rate 0.01 leads to relatively fast and stable convergence. Then we apply the same values on all other datasets. For the first group training, we train \method 100 epochs; for the remaining groups, we train \method 15 epochs as we only need to update the hash embedding table $E$. Note that on every group, we train \method until convergence.

\paragraph{Evaluation metric.} We use accuracy as an evaluation metric. As we sequentially adapt the model on each vocabulary group's training set and test the model on the test set, we have running accuracies on each group.

\begin{figure*}
    \center
    \includegraphics[width=\linewidth]{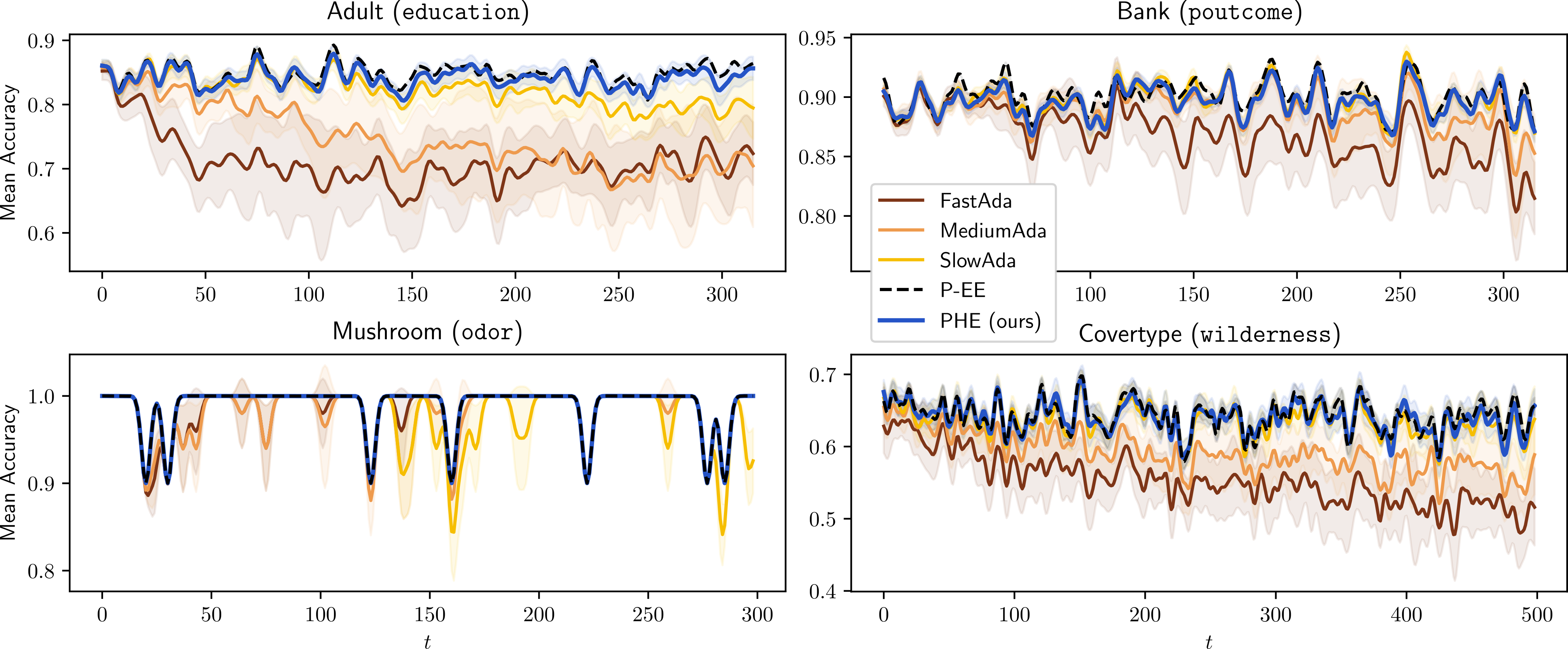}
    \includegraphics[width=\linewidth]{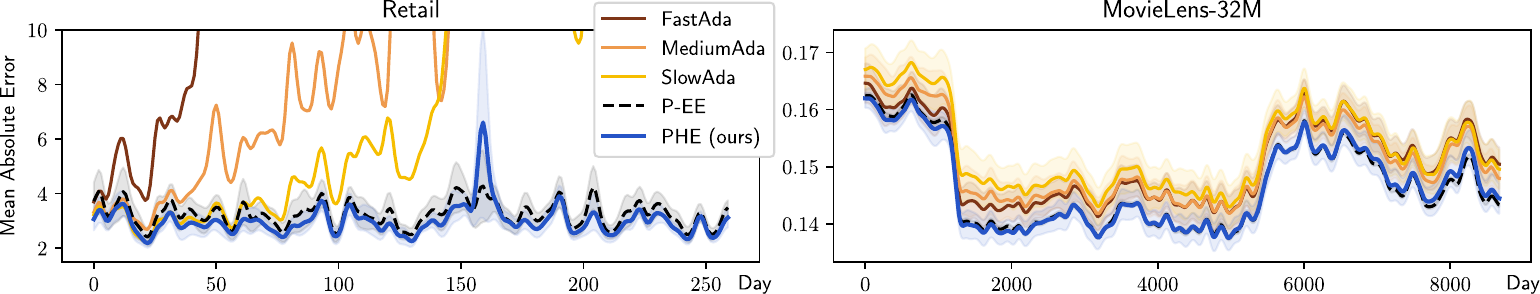}
    
    \caption{Results of online learning on all data. 
    }
    \label{fig:all-tab-online}
\end{figure*}

\paragraph{Additional results.}
We add all datasets' online learning results in \Cref{fig:all-tab-online}.

\subsection{Details for Multi-Task Sequence Modeling Experiments}
\label{sec:app-sequence-mdoel}

\paragraph{Datasets.} We use the Retail dataset\footnote{
  \url{http://archive.ics.uci.edu/dataset/352/online+retail}
} as a multi-task TTD to demonstrate \method.
The dataset involves over 4,000 products indicated by \texttt{StockCode} column and the corresponding sale quantities represented by \texttt{quantity} column with invoice timestamps. We treat \texttt{quantity} as a time series and then track \texttt{quantity} for all 4,000 selling goods over time in a filtering setup. 
Prediction for each piece of product is regarded as one task and there are over 4,000 tasks in total. 
The task is to predict the sales quantity for the product shown in each invoice record given the product's previous sales. 

For the continual learning setting in \Cref{sec:app-continual-learn}, we treat all transactions as occurring at even time intervals. 
For each task, we randomly split the training and testing set with a ratio of 2:1. To get multi-tasks in a dynamic setting, we treat \texttt{StockCode} as the task identifier and evenly partition the products in \texttt{StockCode} into ten disjoint groups where each group involves about 400 goods, i.e., 400 new tasks.
Correspondingly, the original dataset is converted into a task-incremental dataset where each task refers to predicting sale quantities (i.e., taking \texttt{Quantity} column values as $\y$) for one product, indicated by \texttt{StockCode} column. We normalize the \texttt{UnitPrice} column into the range $[0, 1]$ and do not use the \texttt{Description} column. We also drop cancellation transactions that have \texttt{Quantity} values smaller than zero. Therefore, we refer to \texttt{StockCode} as $\u$, \texttt{Quantity} as $\y$, \texttt{UnitPrice} as $\x$, \{\texttt{Country}, \texttt{CustomerId}\} as $\m$, and \texttt{InvoiceDate} as $t$.

\paragraph{Evidence lower bound.} We assume the sales quantity follows Poisson distribution, consequently using the Poisson likelihood. 

\paragraph{Implementation details and hyperparameters.} We also implement a weighted aggregation function $g$ as above in the supervised learning setup. We did not try out different hyperparameter settings and directly set $B=109,K=3,d=20,P=109$ as these values can already support a large vocabulary (of size $P\times B^K$). We apply the same values to both \method and the baselines.

\paragraph{Optimization.} We use Adam stochastic optimization with the same learning rate of 0.005 and the same minibatch size of 128 as in supervised learning experiments. For other hyperparameters of Adam, we apply the default values recommended in the PyTorch framework. For the first task training, we train \method 15 epochs; for the remaining tasks, we train \method 5 epochs. Note that on every task, the epochs used are enough to train \method until convergence.

\paragraph{Evaluation metric.} We also evaluate the performance by the cumulative averages of errors. For each product, we use the first nine observations to predict the 10th observation and measure the absolute error on the 10th observation. Then, the average of all such absolute errors is the performance of this product. Since one group contains about 4,000 products, we further average each product's performance as the group's performance. Specifically, we have a prediction model that has a Poisson likelihood $p(\y_t|\y_{t-9:t-1},\x_{t-9:t},\h_{\m_{t-9:t}}, \h_\u)$. We predict $\hat\y_t=\E[\y_t|\y_{t-9:t-1},\x_{t-9:t},\h_{\m_{t-9:t}}, \h_\u]$ as the mean value and then measure the absolute error between the ground-truth value $|\y_t-\hat\y_t|$. 

For the continual learning setup, after learning group $t$, we can evaluate the performance of all previous and current groups, denoted by $R_{t,\leq t}$. We refer to the cumulative mean absolute error $\bar R_t=\sum_{a=1}^t R_{t,a}/t$ at group $t$ as the performance at $t$. We report $\bar R_t$ as a function of group numbers in \Cref{fig:cum-res-time-tab}. We report $\bar R_T$ after learning the final group $T$ in \Cref{tab:cl-cum-res-dy-tab}.

\paragraph{Additional results.} 
Because we smoothed the results with a 1-D Gaussian filter in the main paper, we provide the first ten days' result without smoothing in \Cref{fig:first-ten-day-online-res}.

\subsection{Details for Recommendation Experiments}
\label{sec:app-exp-recommend}

Beside the first five years, this up-to-date and largest MovieLens dataset
\footnote{
  \url{https://files.grouplens.org/datasets/movielens/ml-32m.zip}
}
have 8688 days (time steps) with possibly no records on some days.
Note after pre-training, all model parameters are fixed except the hash embeddings. Regarding the likelihood function, we assume the rating follows Gaussian distribution.

We randomly split the data into a validation (20\%) and a test set (80\%). We searched the learning rate, batch size, neural network size, and likelihood scale on the validation set and reported final results on the test set. 
PHE, EE, and FastAda train the hash embeddings for 5 epochs per time step while MediumAda trains 2 epochs and SlowAda trains 1 epoch. 

We also use the mean absolute error as the evaluation metric. 

\begin{figure*}
    \center
    \includegraphics[width=0.49\linewidth]{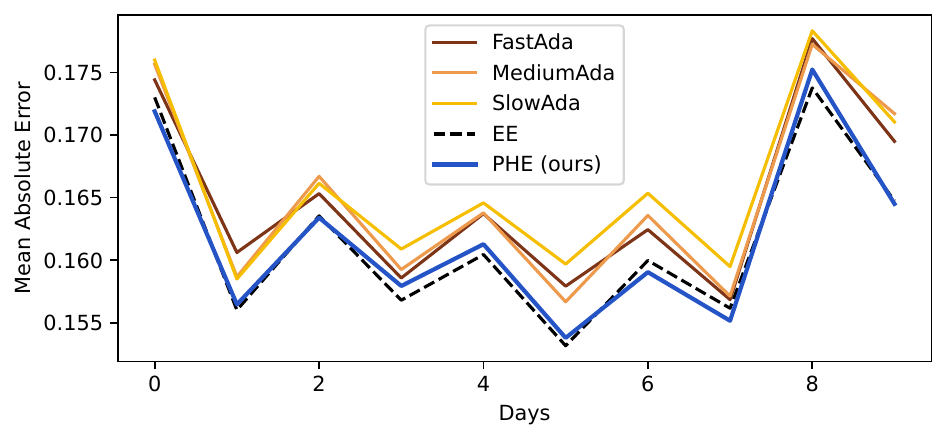}
    \includegraphics[width=0.49\linewidth]{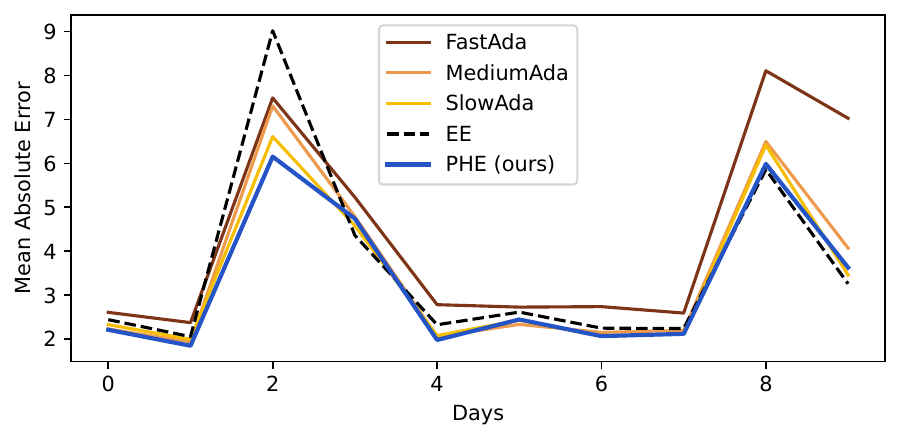}
    \vspace{-0.5em}
    \caption{First ten day results of data-streaming movie recommendation and sales quantity sequence modeling. 
    }
    \label{fig:first-ten-day-online-res}
\end{figure*}

\paragraph{Results.} We plot the online learning results in \Cref{fig:all-tab-online}. The curves in are smoothed with a 1-D Gaussian filter. The initial performance gap on Day 0 is an artifact of smoothing, in fact, all methods have similar performance initially (see \Cref{fig:first-ten-day-online-res} in \Cref{sec:app-exp-recommend}). An interesting observation: MovieLens made two major changes in their rating system around 2003 and 2014 \url{https://grouplens.org/blog/movielens-datasets-context-and-history/}{(link)}, which is reflected in our results -- two sharp changes in the online learning plot.

\subsection{Additional Results}
\label{sec:app-add-exp}

\subsubsection{More Motivation Examples}
\label{sec:app-motivatino-eg}
\begin{figure}
    \begin{center}
           \includegraphics[width=0.47\textwidth]{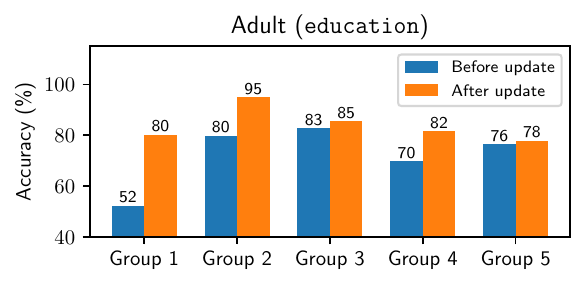} 
           \includegraphics[width=0.47\textwidth]{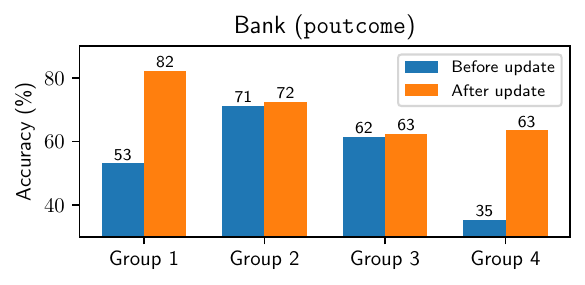}
           \includegraphics[width=0.47\textwidth]{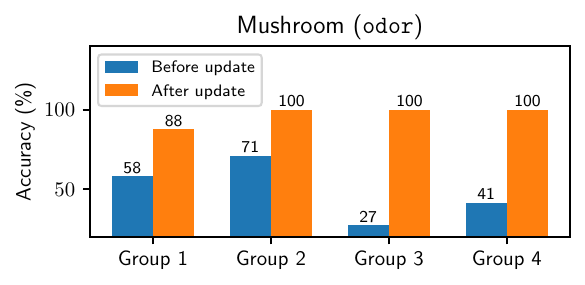}
           \includegraphics[width=0.47\textwidth]{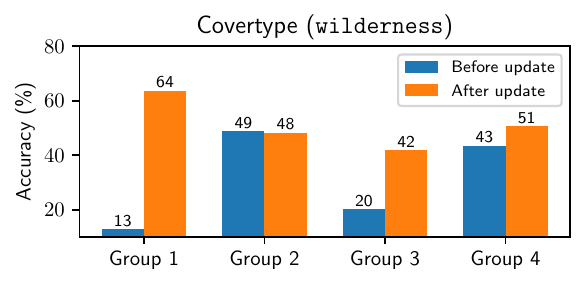} 
        \caption{Adult dataset is randomly split into disjoint groups based on the \texttt{education} column. Groups arrive sequentially. We report results before and after the updates on the hash embeddings for each group to motivate the need to incorporate new groups into the model. Results are averaged on five independent runs with different random parameter initializations.
         }
         \label{fig:adult-per-group}
    \end{center}
\end{figure}

\begin{figure}
\tiny
\begin{center}
\begin{minipage}[c]{0.5\linewidth}
Adult
\begin{itemize}
    \item[Group 1] (\texttt{Preschool, 5th-6th, Bachelors})
    \item[Group 2] (\texttt{10th, 11th, 12th})
    \item[Group 3] (\texttt{7th-8th, HS-grad, Prof-school})
    \item[Group 4] (\texttt{9th, Assoc-voc, Doctorate})
    \item[Group 5] (\texttt{1st-4th, Masters, Some-college, Assoc-acdm})
\end{itemize}

Mushroom
\begin{itemize}
    \item[Group 1] (\texttt{Musty, None})
    \item[Group 2] (\texttt{Anise, Almond})
    \item[Group 3] (\texttt{Spicy, Creosote})
    \item[Group 4] (\texttt{Foul, Fishy, Pungent})
\end{itemize}

CoverType
\begin{itemize}
    \item[Group 1] (\texttt{A1})
    \item[Group 2] (\texttt{A3})
    \item[Group 3] (\texttt{A4})
    \item[Group 4] (\texttt{A2})
\end{itemize}

Bank
\begin{itemize}
    \item[Group 1] (\texttt{Unknown})
    \item[Group 2] (\texttt{Failure})
    \item[Group 3] (\texttt{Other})
    \item[Group 4] (\texttt{Successs})
\end{itemize}
\end{minipage}
\end{center}

    \caption{Group information for continual classification tasks.}
    \label{fig:group-info}
\end{figure}

We report additional results in \Cref{fig:adult-per-group} as a complement to \Cref{fig:group-update-perf} in the main paper. \Cref{fig:adult-per-group} provides more evidence for the motivation of our work. For tabular data in a dynamic setting, not including the newly created categorical feature values in the prediction model will lead to a performance drop. Therefore, an efficient way to incorporate the new categorical features is necessary to maintain the efficacy of a prediction model. The ``After update'' performance in the plots demonstrates PHE is desirable for adapting to the new features. The splitting details are in \Cref{sec:app-classification,sec:app-sequence-mdoel,fig:group-info}.

\subsubsection{Memory Efficiency}
\label{sec:app-memory}

\begin{table*}[htb]
\caption{
Number of parameters in the embedding module. 
Ratios are computed by dividing PHE by P-EE. The results show that PHE consumes as little as 2\% of the number of parameters (i.e., hardware memory) of P-EE, demonstrating the memory-efficiency benefit of PHE. (See details in \Cref{sec:app-memory}.)%
} 
\label{tab:memory-cost}
\small
\centering
\resizebox{0.6\textwidth}{!}{
\begin{tabular}{l|cccccc}
\toprule
  &Adult &Bank &Covertype &Mushroom  &Retail &MovieLens-32M \\
\midrule
PHE (ours) 
    &346 
    &346 
    &346 
    &56 
    &5014 
    &460414\\
P-EE 
    &3920 
    &1760 
    &1760 
    &90 
    &332760 
    &11541320\\

\midrule
Compression Ratio &0.09 &0.2 &0.2 &0.62 &0.02 &0.04\\
\bottomrule
\end{tabular}
}
\vspace{-0.5em}
\end{table*}

Memory efficiency of PHE can be seen from the number of parameters in the embedding module, which we summarized for both PHE and P-EE in \Cref{tab:memory-cost}.
Note that P-EE sets the performance upper bound but its size scales linearly with the vocabulary size. 
The fact that PHE on all datasets achieves the same performance as P-EE illustrates PHE's impressive memory efficiency, especially considering PHE only consumes as low as $2\%$ memory of P-EE.
Besides, being a unified embedding where all categorical columns share the same embedding table \citep{coleman2023unified}, PHE is compatible with modern hardware and can benefit from the hardware acceleration.

We multiply each number by two because every parameter has its mean and variance. 20 is due to each embedding has 20 dimensions. 
For PHE, refer to implementation details (\Cref{app:exp}) for the number of parameters ($B\times d + P\times K$). 
We compute the P-EE parameter size by $V\times d$ where $V$ is the vocabulary size.

\subsubsection{Adaptation and Forgetting analysis}
\label{sec:app-adapt-forget}
\begin{figure*}[t!]
    \center
    \includegraphics[width=0.95\linewidth]{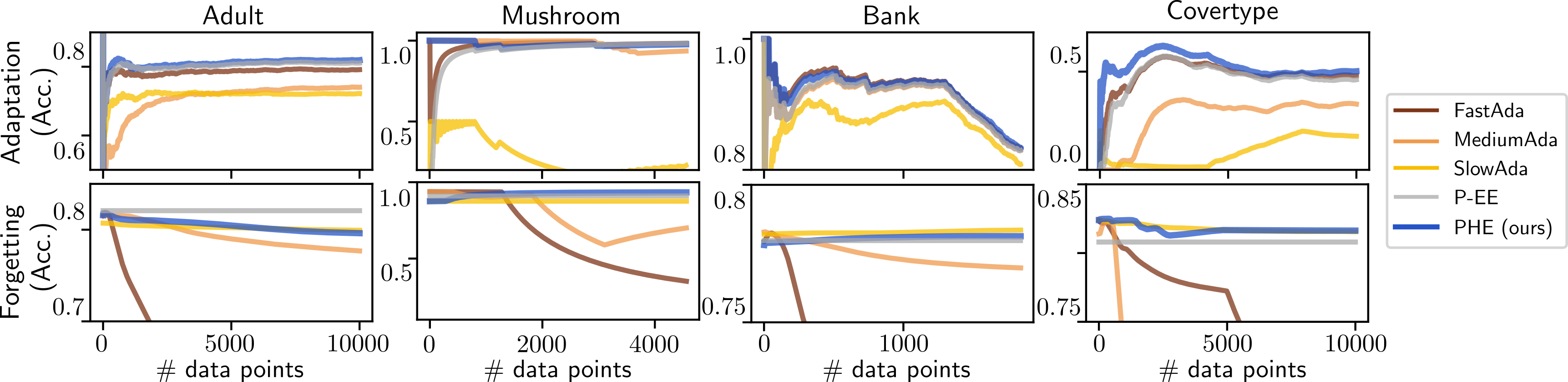}
    \vspace{-0.5em}
    \caption{Comparision of online learning methods' adaptation and forgetting in a streaming online setup. 
    Our PHE achieves similar performance with the collision-free P-EE on both metrics. 
    Notably, SlowAda forgets the least but is slow in adaptation; FastAda is in the opposite regime.
    }
    \label{fig:streaming}
\end{figure*}

We designed experiments to specifically measure the adaptation to new data and forgetting of old data. We split the data into two disjoint groups based on a random partition of one column's vocabulary. 
The model was initialized using the first group and online updated on the second group whose items are unseen in initialization.
We let the data arrive one at a time. 
Adaptation is measured by the cumulative predictive accuracy of new datum and the forgetting by the accuracy of the first group's test data. 
Results in \Cref{fig:streaming} show that our PHE has almost the best adaptation and forgetting performance on all four datasets. 
The P-EE while does not suffer forgetting, its adaptation to new categories is slow as each new embedding is initialized at random. 

Regarding baselines, SlowAda uses a small learning rate (1e-4); MediumAda uses a medium learning rate (1e-3); FastAda uses a large learning rate (1e-2). 

In \Cref{fig:streaming}, we compare on all four classification datasets used in the paper, our PHE against the four baselines. We observe from \Cref{fig:streaming} that the SlowAda baseline with smaller LR (1e-4) leads to slower forgetting at the cost of slower adaptation, while larger LR (1e-2) has faster adaptation at the cost of faster forgetting (FastAda). Thus a data-stream dependent LR is needed for deterministic hash embeddings to trade off adaptation and forgetting. In contrast, our PHE has almost the best adaptation and forgetting performance on all four datasets due to the regularization from the posteriors. The EE while does not suffer forgetting as each category has a separate row in the embedding table, its adaptation to new categories is slow as each new embedding is initialized at random. 

\subsubsection{Continual learning}
\label{sec:app-continual-learn}

\begin{figure}[t!]
    \centering
           \includegraphics[width=0.98\textwidth]{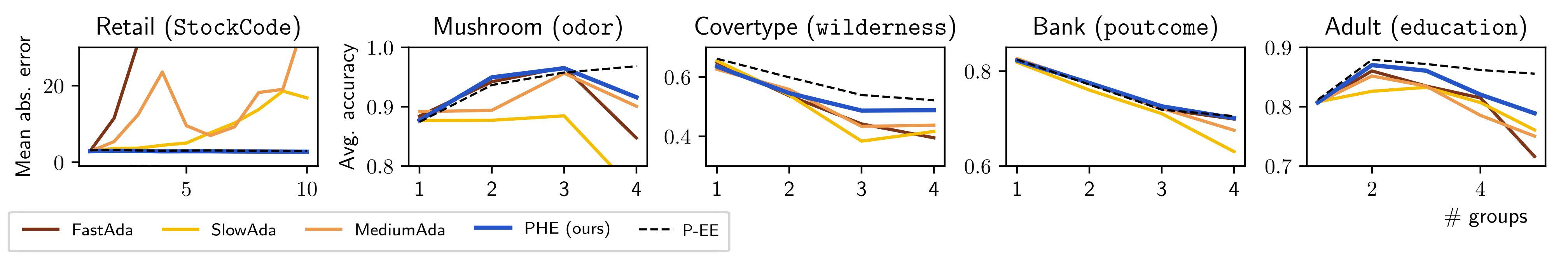}
        \vspace{-0.5em}
        \caption{Cumulative average results in continual learning. Column names in the parentheses are the ones made to have changing vocabulary and used to split groups. 
        PHE is closest to the performance upper-bound P-EE.
            }
        \label{fig:cum-res-time-tab}
\end{figure}

\begin{table}[t]
\caption{Performance on classification and sequence modelling tasks. Adult, Bank, Mushroom, and Covertype are classification tasks and thus evaluated by average accuracy, which is larger the better. Retail is a regression task and we use the metric mean absolute error, lower the better.
}
\label{tab:cl-cum-res-dy-tab}
\centering
\small
\begin{tabular}{lccccc}
\toprule
  &SlowAda &MediumAda &FastAda &P-EE (collision-free) &PHE (ours)\\
\midrule
Adult &76.1$\pm$1.8 &75.0$\pm$4.7 &71.6$\pm$3.1 &85.6$\pm$0.1 &78.9$\pm$3.0\\
Bank &63.0$\pm$4.0 &67.5$\pm$4.5 &69.9$\pm$1.2 &70.5$\pm$0.7 &70.1$\pm$1.4\\
Mushroom &75.5$\pm$7.6 &90.1$\pm$8.6 &84.7$\pm$12.3 &96.8$\pm$0.0 &91.6$\pm$7.6\\
Covertype &41.7$\pm$4.0 &43.8$\pm$5.7 &39.5$\pm$5.1 &52.2$\pm$1.1 &48.8$\pm$2.3\\
Retail &16.8$\pm$17.6 &38.9$\pm$50.9 &- &2.92$\pm$0.16 &2.73$\pm$0.23\\
\bottomrule
\end{tabular}
\end{table}

We also investigated classification and sequence modeling in the continual learning setup~\citep{kirkpatrick2017overcoming}, we split the dataset into disjoint groups based on a random partition of a selected column's vocabulary, assuming data distribution differs conditioned on each partition.
This is similar to \Cref{sec:app-motivatino-eg}.
We then sequentially update the embeddings on each group's training data. 
After each group training, we evaluated the model performance on all previously seen groups' test data. The splitting details are in \Cref{sec:app-classification,sec:app-sequence-mdoel,fig:group-info}.
While data-streaming setup aims to have good performance on the latest task, the goal of continual learning is to perform well on all groups after sequential training.
\Cref{fig:cum-res-time-tab} and \Cref{tab:cl-cum-res-dy-tab} summarizes the results. Our PHE has the top performance among hash embedding methods.

\subsubsection{Ablation studies}
\label{sec:app-ablate}
\begin{table}[t]
\caption{Comparison between updating all categorical columns' embeddings, only updating incremental columns' embeddings, and updating all model parameters. We used collision-free expandable embeddings in the experiments. The first two updating protocols have little difference but updating all parameters sometimes result in performance deterioration, possibly due to catastrophic forgetting in the network weights.
} 
\label{tab:ablation-incre-only-update}
\begin{center}
\resizebox{\textwidth}{!}{
\begin{tabular}{lccccc}
\toprule
  &Adult (Acc.) &Bank (Acc.) &Mushroom (Acc.) &Covertype (Acc.) &Retail (Err.) \\
\midrule
Update all columns embeddings
    &84.7$\pm$0.0 
    &90.0$\pm$0.0 
    &98.8$\pm$0.0 
    &64.1$\pm$0.0
    &3.4$\pm$0.3\\
Update incremental columns embeddings (in use)
    &84.8$\pm$0.0 
    &90.1$\pm$0.0 
    &98.8$\pm$0.0 
    &64.0$\pm$0.4
    &3.2$\pm$0.4\\
Update all model parameters
    &83.3$\pm$0.1
    &89.5$\pm$0.0 
    &98.8$\pm$0.0 
    &64.0$\pm$0.1
    &287.9$\pm$125.5\\
\bottomrule
\end{tabular}
}
\end{center}
\end{table}
\paragraph{Variants of updating protocols.}
We provided evidence on our updating protocols in \Cref{tab:ablation-incre-only-update}, showing updating incremental column's embeddings as well as fixing other parameters has the best performance. 
\Cref{tab:ablation-incre-only-update} presents the accuracy of multiple updating schemes, justifying this updating protocol in use achieves both high accuracy and computational efficiency. 

\paragraph{The impact of potential hash collisions and the mitigation measures.}

We experimented on the large Retail dataset under the continual learning setup as in \Cref{sec:app-continual-learn}. We varied the hyperparameters bucket size B and the number of hash functions K to control the potential number of hash collisions. In particular, we varied one hyperparameter when fixing the other.

We repeated each experiment five times with different random seeds. The tables below show the mean absolute errors (the lower the better) with standard deviation under each hyperparameter setting. In the first table, we varied bucket size B while fixing the number of hash functions to be K=2. In the second table, we fixed the bucket size B to 109, which is the same as in the paper, and changed the number of hash functions. The collision probability increases from right to left for both tables. The results in the first table show the more likely a hash collision, the more unstable the model performance. However, the deterioration is slow, showing the method's robustness to potential hash collisions and various hyperparameter settings. In the second table, although increasing K reduces the probability of hash collisions, increasing K also increases the number of effective parameters (related to model complexity) to fit in the model. It thereby increases the variance of the predictive performance. Thus, we recommend choosing a small K (such as 2-3) that trades off both hash-collision and predictive performance variance. Note when K=1, the hash collision will cause two items to have exactly the same resulting hash embeddings, leading to a high variance among all settings. We will add these results to the ablation section in the revised paper.

\begin{center}
Ablation study on bucket size B\\
{\small 
\begin{tabular}{cccc}
\toprule
  B=40,K=2 &B=60,K=2 &B=80,K=2 &B=109,K=2 \\
\midrule
2.83$\pm$0.23
&2.65$\pm$0.16
&2.56$\pm$0.10
&2.58$\pm$0.09\\
\bottomrule
\end{tabular}
}

Ablation study on the number of hash functions K\\
{\small 
\begin{tabular}{ccccc}
\toprule
  B=109,K=1 &B=109,K=2 &B=109,K=3 &B=109,K=4  &B=109,K=5\\
\midrule
2.66$\pm$0.34
&2.58$\pm$0.09
&2.63$\pm$0.16
&2.78$\pm$0.18
&2.76$\pm$0.13\\
\bottomrule
\end{tabular}
}
\end{center}

\textit{Remedy.} We use the standard trick of multiple independent hash functions to reduce the collision probability of two unique items. As is standard in universal hashing [Carter and Wegman, 1997], the probability of collision with all K hash functions each hashing into B buckets is proportional to 
 (see section 3.3). Collision of hash values could happen among popular, important categories. To address this issue, we can select the desired hash functions that avoid important collisions before applying the hash functions. In addition, users come and go fast, and collisions may become unimportant over time.

\paragraph{Double memory of Ada baselines.} In \Cref{tab:cum-res-dy-tab}, Ada baselines do not have the same number (actually half) of parameters as PHE because PHE maintains a pair of mean and variance parameters for each embedding. We conducted another set of experiments on the Ada baselines using the same settings as in Tab. 1, except that we doubled the size of embedding tables, i.e., $B\rightarrow2B$ and $P\rightarrow2P$. This way, both PHE and the deterministic Ada baselines have the same number of parameters, but Ada has much lower collision rates. We report the results in the table below. It shows that: 1) PHE is still the top performer; 2) the average performance gap between the Ada baselines and PHE narrows; 3) no single Ada baseline works well across all datasets; and 4) PHE has smaller performance variation. 
\begin{center}
Results of double-memory Ada baselines\\
\begin{tabular}{lcccc}
\toprule
  &SlowAda &MediumAda &FastAda &PHE (ours)\\
\midrule
Adult &83.6 $\pm$ 
0.9	&79.9 $\pm$
 2.2	&76.1 $\pm$
 2.5	&\textbf{84.1 $\pm$ 
 0.2}\\
Bank &\textbf{89.7 $\pm$
 0.1} &89.4 $\pm$
 0.3	&87.7 $\pm$
 1.6	&\textbf{89.6 $\pm$
 0.0}\\
Mushroom &97.9 $\pm$
 1.5	&98.4 $\pm$
 0.5	&98.5 $\pm$
 0.2	&\textbf{98.8 $\pm$
 0.0}\\
Covertype &\textbf{64.2 $\pm$
 0.6}	&61.6 $\pm$
 1.0	&55.8 $\pm$
 3.2	&\textbf{64.3 $\pm$
 0.2}\\
Retail &6.4 $\pm$
 0.6	&8.2 $\pm$
 2.3	&-	&\textbf{3.0 $\pm$
 0.2}\\
MovieLens &15.4 $\pm$
 0.1	&15.1 $\pm$
 0.0	&15.2 $\pm$
 0.1	&\textbf{14.7 $\pm$
 0.0}\\
\bottomrule
\end{tabular}
\end{center}

\end{document}